\documentclass[sigconf]{acmart}

\usepackage{booktabs} % For formal tables

\usepackage{multirow}
\usepackage{multicol}
\usepackage{soul}
\usepackage{amsthm}
\usepackage{mathtools, cuted}

\usepackage{algorithm}
\usepackage[noend]{algpseudocode}

\usepackage{tikz}
\usetikzlibrary{shapes.geometric, arrows}
\tikzstyle{normal} = [rectangle, rounded corners, minimum width=1.0cm, minimum height=0.8cm,text centered, draw=black]
\tikzstyle{arrow} = [thick,->,>=stealth]

\newtheorem{remark}{Remark}

\setcopyright{rightsretained}
\acmDOI{10.475/123_4}

\acmISBN{123-4567-24-567/08/06}

\acmConference[KDD'18]{ACM Woodstock conference}{19 - 23 August 2018}{London UK}
\acmYear{2018}
\copyrightyear{2018}

\acmArticle{4}
\acmPrice{15.00}

\begin{document}
\title{Graph-Based Deep Modeling and Real Time Forecasting of Sparse Spatio-Temporal Data}

%\titlenote{Produces the permission block, and copyright information}
%\subtitle{Extended Abstract}
%\subtitlenote{The full version of the author's guide is available as
%  \texttt{acmart.pdf} document}

%\author{Bao Wang, Xiyang Luo, Fangbo Zhang, \newline Andrea L. Bertozzi}
\author{Bao Wang, Xiyang Luo, Fangbo Zhang}\authornote{Bao Wang, Xiyang Luo, and Fangbo Zhang contributed equally.}
\author{Baichuan Yuan, Andrea L. Bertozzi}
%\authornote{Bao Wang, Xiyang Luo, and Fangbo Zhang are equally contributed.}
\orcid{1234-5678-9012}
\affiliation{
  %\institution{Department of Mathematics, UCLA, Los Angeles, CA}
  \institution{Dept of Math, UCLA, Los Angeles, CA}
  %\streetaddress{P.O. Box 951555}
  \postcode{90095-1555}
}
\email{wangbaonj@gmail.com, xylmath@gmail.com, fb.zhangsjtu@gmail.com }
\email{ybcmath@gmail.com, bertozzi@math.ucla.edu}

\author{P. Jeffrey Brantingham}
\orcid{1234-5678-9012}
\affiliation{
  %\institution{Department of Anthropology, UCLA, Los Angeles, CA}
  \institution{Dept of Anthropology, UCLA, Los Angeles, CA}
  %\streetaddress{P.O. Box 951555}
  \postcode{90095-1555}
}
\email{branting@ucla.edu}

%\authornote{The secretary disavows any knowledge of this author's actions.}
%\affiliation{%
%  \institution{Institute for Clarity in Documentation}
%  \streetaddress{P.O. Box 1212}
%\email{webmaster@marysville-ohio.com}
%
%\author{Lars Th{\o}rv{\"a}ld}
%\authornote{This author is the
%  one who did all the really hard work.}
%\affiliation{%
%  \institution{The Th{\o}rv{\"a}ld Group}
%  \streetaddress{1 Th{\o}rv{\"a}ld Circle}
%\email{larst@affiliation.org}
%\affiliation{%
%  \institution{Inria Paris-Rocquencourt}
%\affiliation{%
% \institution{Rajiv Gandhi University}
%\affiliation{%
%  \institution{Tsinghua University}
%  \streetaddress{30 Shuangqing Rd}
%  \country{China}
%}

%\affiliation{%
%  \institution{Palmer Research Laboratories}
%  \streetaddress{8600 Datapoint Drive}

%\affiliation{\institution{The Th{\o}rv{\"a}ld Group}}
%\email{jsmith@affiliation.org}

%\author{Julius P.~Kumquat}
%\affiliation{\institution{The Kumquat Consortium}}
%\email{jpkumquat@consortium.net}

\renewcommand{\shortauthors}{B. Wang et al.}

\begin{abstract}
We present a generic framework for spatio-temporal (ST) data modeling, analysis, and forecasting, with a special focus on data that is sparse
in both space and time. Our multi-scaled framework is a seamless coupling of two major components:
%\xmod{\st{a stochastic process model} a self exciting point process that}
a self-exciting point process that
models the macroscale statistical behaviors
of the ST data and a graph structured recurrent neural network (GSRNN) to discover the
microscale patterns of the ST data on the inferred graph. This novel deep neural network (DNN) %structure inherits good generalization abilities
%from the DNN and
incorporates the real time interactions of the graph nodes to enable more accurate real time forecasting.
The effectiveness of our method is demonstrated on both crime and traffic forecasting.
\end{abstract}



%\ccsdesc[500]{Computer systems organization~Embedded systems}


\maketitle

\section{Introduction}
Accurate spatio-temporal (ST) data forecasting is one of the central tasks for artificial intelligence with many practical applications. For instance, accurate crime forecasting can be used to prevent criminal behavior, and forecasting traffic is of great importance for urban transportation system. Forecasting the ST distribution effectively is quite challenging, especially at hourly or even finer temporal scales in micro-geographic regions. The task becomes even harder when the data is spatially and/or temporally sparse. There are many recent efforts devoted to quantitative study of ST data, both from the perspective of statistical modeling of macro-scale properties and deep learning based approximation of micro-scale phenomena. We briefly mention a few relevant works. %\cite{Mohler:2011JASA}, %\cite{Mohler:2011JASA, Short:2014DCDSB, Short:2010SIAMADS, Short:2010PNAS},
Mohler et al pioneered the use of the Hawkes process (HP) to predict crime.  Recent field trials \cite{Mohler:2011JASA} show these models can outperform crime analysts, and are now used in commercial software deployed in over 50 municipalities worldwide. In \cite{Kang:2017Crime}, the authors utilized a convolutional neural network (CNN) to extract the features from the historical crime data, and then used a support vector machine (SVM) to classify whether there will be crime or not at the next time slot. Zhang et al \cite{Junbo:2017} create an ensemble of residual networks \cite{He:2016ResNet}, named ST-ResNet, to study and predict traffic flow. Additional applications include \cite{Jain:2016} who use the ST graph to represent human environment interaction, and proposed a structured recurrent neural network (RNN) for semantic analysis and motion reasoning. The combination of video frame-wise forecasting and optical flow interpolation allows for the forecasting of the dynamical process of the robotics motion \cite{Holden:2017}. RNNs have also been combined with point processes to study taxi and other data \cite{Du:2016}.

This paper builds on our previous work \cite{BaoWang:2017DL1} in which we applied ST-ResNet, along with data augmentation techniques, to forecast crime on a small spatial scale in real time. We further showed that the ST-ResNet can be quantized for crime forecasting with only a negligible precision reduction \cite{BaoWang:2017DL2}. Moreover, ST data forecasting also has wide applications in computer vision \cite{LeCun:2015Nature, Hochreiter:1997NC, Jain:2016, Holden:2017, Li:2016GatedGraphNN}.
%In the previous CNN based approaches for crime or traffic forecasting,
Many previous CNN-based approaches for ST forecasting map the spatial distribution to a rectangular box partitioned with a rectangular grid.
The data at a certain timescale is represented by a histogram on the grid.
Finally, a CNN is used to predict the future histogram. This prototype is sub-optimal
from two aspects. First, the geometry of a city is usually highly irregular, resulting in the city's configuration taking up only a small portion of its bounding box.
This introduces unnecessary redundancy into the algorithm. Second, the spatial sparsity can be exacerbated by the
spatial grid structure. Directly applying a CNN to fit the extreme sparse data will lead to all zero weights
due to the weight sharing of CNNs \cite{BaoWang:2017DL2}. This can be alleviated by using spatial super-resolution\cite{BaoWang:2017DL2}, with
increased computational cost. Moreover this lattice based data representation omits geographical information and spatial correlation within the data itself.

\begin{figure}% code for the flow chart
    \centering
    \begin{tikzpicture}[node distance=1cm]
    \node (data) [normal] {Data};
    \node (graph) [normal, right of=data, xshift=1.35cm] {Graph};
    \node (dnn) [normal, right of=graph, xshift=1.35cm] {DNN};
    \node (output) [normal, right of=dnn, xshift=1.35cm] {Output};
    \draw [arrow] (data) -- (graph);
    \draw [arrow] (graph) -- (dnn);
    \draw [arrow] (dnn) -- (output);
    \end{tikzpicture}
    \caption{Flow chart of the algorithm.}
    \label{fig:flow-chart}
\end{figure}
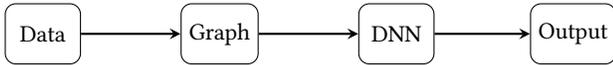

% Spatial partition
In this work, we develop a generic framework to model sparse and unstructured ST data.
Compared to previous ad-hoc spatial partitioning, we introduce an ST weighted graph (STWG) to represent the data, which automatically solves the issue caused by spatial sparsity. This STWG carries the spatial cohesion and temporal
evolution of the data in different spatial regions over time. We infer the STWG by solving a statistical
inference problem. For crime forecasting, we associate each graph node with a time series of crime intensity
in a zip code region, where each zip code is a node of the graph. As is shown in \cite{Mohler:2011JASA},
the crime occurrence can be modeled by a multivariate Hawkes process (MHP), where the self and mutual-exciting rates determines the connectivity
and weights of the STWG. To reduce the complexity of the model, we enforce the graph connectivity to be sparse. To this end, we add an
additional $L_1$ regularizer to the maximal likelihood function of MHP. The inferred STWG incorporates the macroscale evolution of the
crime time series over space and time, and is much more flexible than the lattice representation. To perform micro-scale
forecasting of the ST data, we build a scalable graph structured RNN (GSRNN) on the inferred graph based on the structural-RNN (SRNN) architecture \cite{Jain:2016}.
Our DNN is built by arranging RNNs in a feed-forward manner: We first assign a cascaded long short-term memory (LSTM) (we will explain this in the following paragraph) to fit the time series on each node of the graph. Simultaneously, we associate each edge of the graph with a cascaded LSTM that receives the output
from neighboring nodes along with the weights learned from the Hawkes process. Then we feed the tensors learned by these edge LSTMs to their
terminal nodes. This arrangement of edge and node LSTMs gives a native feed-forward structure that is different from the classical multilayer perceptron. A neuron is the basic building block of the latter, while our GSRNN is built with LSTMs as basic units.
The STWG representation together with the feed-forward arranged LSTMs build the framework for ST data forecasting.
The flowchart of our framework is shown in Fig. \ref{fig:flow-chart}.

% Statement the contribution.
Our contribution is summarized as follows:
\begin{itemize}
\item We employ a compact STWG to represent the ST sparse unstructured data, which automatically encodes important statistical properties of the data.
\item We propose a simple data augmentation scheme to allow DNN to approximate the temporally sparse data.
\item We generalize the SRNN\cite{Jain:2016} to be bi-directional, and apply a weighted average pooling which is
more suitable for ST data forecasting.
\item We achieve remarkable performance on real time crime and traffic forecasting at a fine-grained scale.
\end{itemize}

%This paper is organized as follows: In section \ref{ProblemStatement}, we rigorously formulate the problem of ST sparse unstructured data modeling and forecasting.
In section \ref{Datset}, we describe the datasets used in this work, including data acquisition, preprocessing, spectral and simple statistical analysis. In section \ref{Algorithms}, we present the pipeline for general ST data forecasting, which contains STWG inference, DNN approximation of the historical signals, and a data augmentation scheme. Numerical experiments on the crime and traffic forecasting tasks are demonstrated in sections \ref{ExperimentsCrime} and \ref{ExperimentsTraffic}, respectively. The concluding remarks and future directions are discussed in section \ref{Conclusion}.

\section{Dataset Description and Simple Analysis}\label{Datset}
We study two different datasets: the crime and traffic data. The former is more irregular in space and time, and hence is much more challenging to study.
In this section we describe these datasets and introduce some preliminary analyses.

\subsection{Crime Dataset}
\subsubsection{Data Collection and Preprocessing}
We consider crime data in Chicago (CHI) and Los Angeles (LA). In our framework, %\xnote{prefer pipeline over pipeline? },
historical crime and weather data are the key ingredients. Holiday information, which is easy to obtain, is also included. The time intervals studied are 1/1/2015-12/31/2015 for CHI and 1/1/2014-12/31/2015 for LA, with a time resolution of one hour. Here we provide a brief description of the acquisition of these two critical datasets.

\paragraph{{\bf Weather Data}}
We collect the weather data from the Weather Underground data base\footnote{\url{https://www.wunderground.com/}}
%(\url{https://www.wunderground.com/}),
through a simple web crawler.
%For the web data, since the format varies day by day\FB{, s}pecial attention should be paid to get the correct data\FB{(should we mention this?)}.
We select temperature, wind speed, and special events, including fog, snow, rain, thunderstorm for our weather features.
%Since this paper focuses on predicting crime on an hourly basis, the weather data is averaged if they are collected within the same hour. Linear interpolation is used to fill missing values for some time slots.
All data is co-registered in time to the hour.
\paragraph{{\bf Crime Data}}
The CHI crime data is downloaded from the \href{https://data.cityofchicago.org/Public-Safety/Crimes-2001-to-present/ijzp-q8t2}{City of Chicago open data portal}. The LA data is provided by the LA Police Department (LAPD).
%\xnote{(This should to go footnote: A special thank goes to LAPD for providing us the historical data.)}
Compared to CHI data, the LA crime data is sparser and more irregular in space. We first map the crime data to the corresponding postal code using QGIS software \cite{QGIS_software}. A few crimes in CHI (less than $0.02\%$) cannot be mapped to the correct postal code region, and we simply discard these events. For the sake of simplicity, we only consider zip code regions with more than 1000 crime events over the full time period. This filtering criterion retains over 95 percent of crimes for both cities, leaving us with 96 postal code regions in LA and 50 regions in CHI.

\subsubsection{Spectrum of the Crime Time Series}
%Figure \ref{CrimeTimeSeries} plots the hourly crime intensities over the entire LA and CHI areas, and over randomly selected postal code regions of the two cities. From %appearances, these time series are quite noisy with very little predictable signal. However, as is shown in Fig. \ref{Spectrum-CrimeTimeSeries}, it is easy to see that the %spectrums of all four time series are peaked at 24, which indicates these time series all have periodic signal with period equal to one day. Additional tests show this is %true for all regions in LA and Chicago. Furthermore, the spectrums of the time series of the selected zipcode regions are noisier than that of the entire city.

\begin{figure}
\centering
\includegraphics[width=0.45\columnwidth]{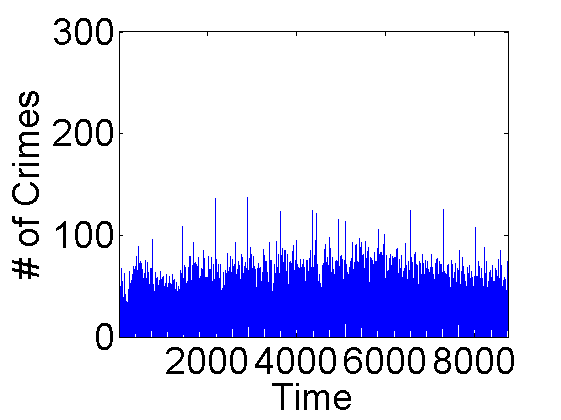}
\includegraphics[width=0.45\columnwidth]{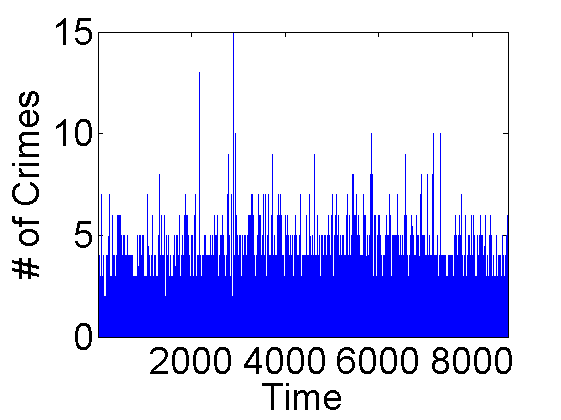}
\caption{Example plots of the hourly crime intensities for the entire 2015 CHI (left) and the 2015 60620 zip code (right).}
\label{CrimeTimeSeries}
\end{figure}

\begin{figure}
\includegraphics[width=0.85\columnwidth]{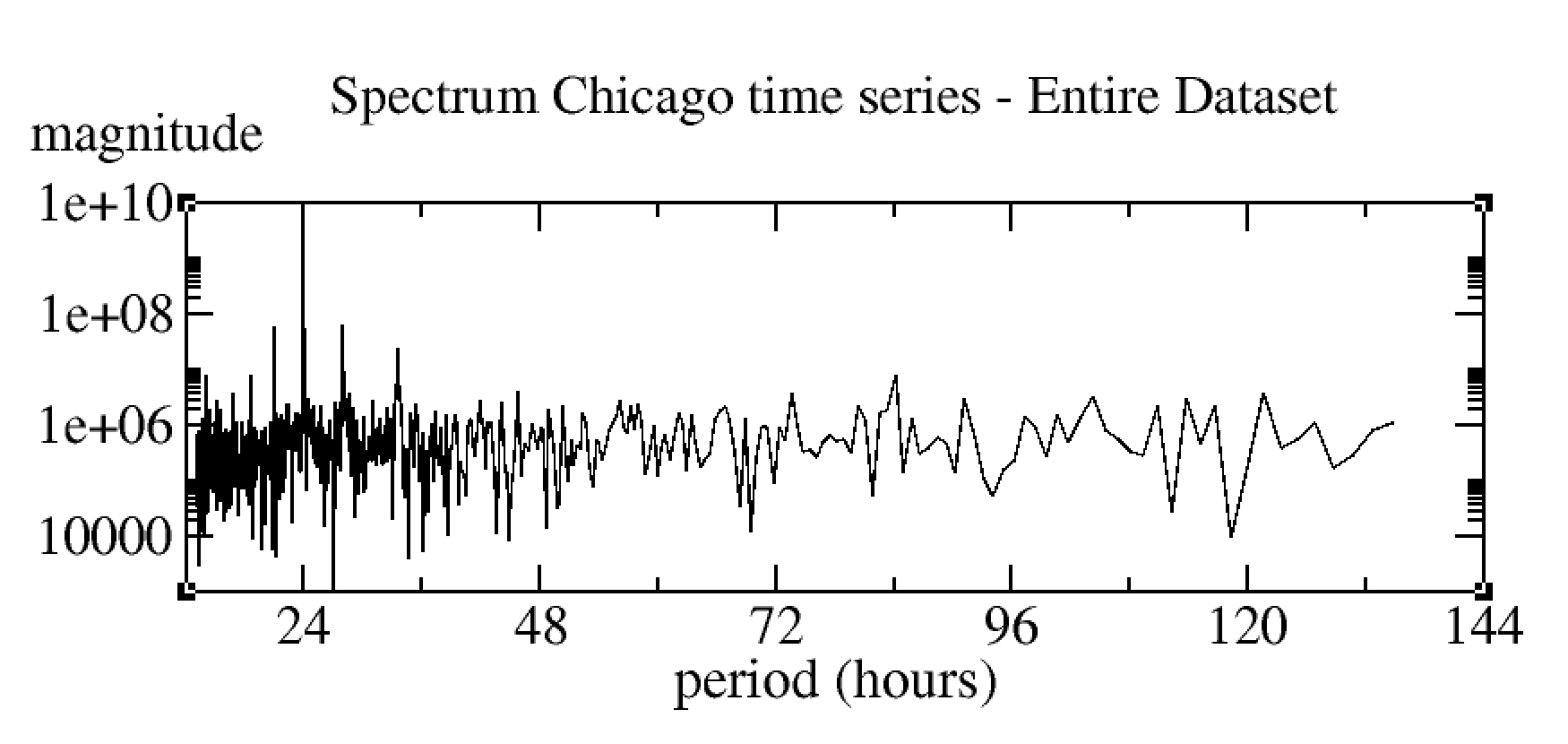}
\includegraphics[width=0.85\columnwidth]{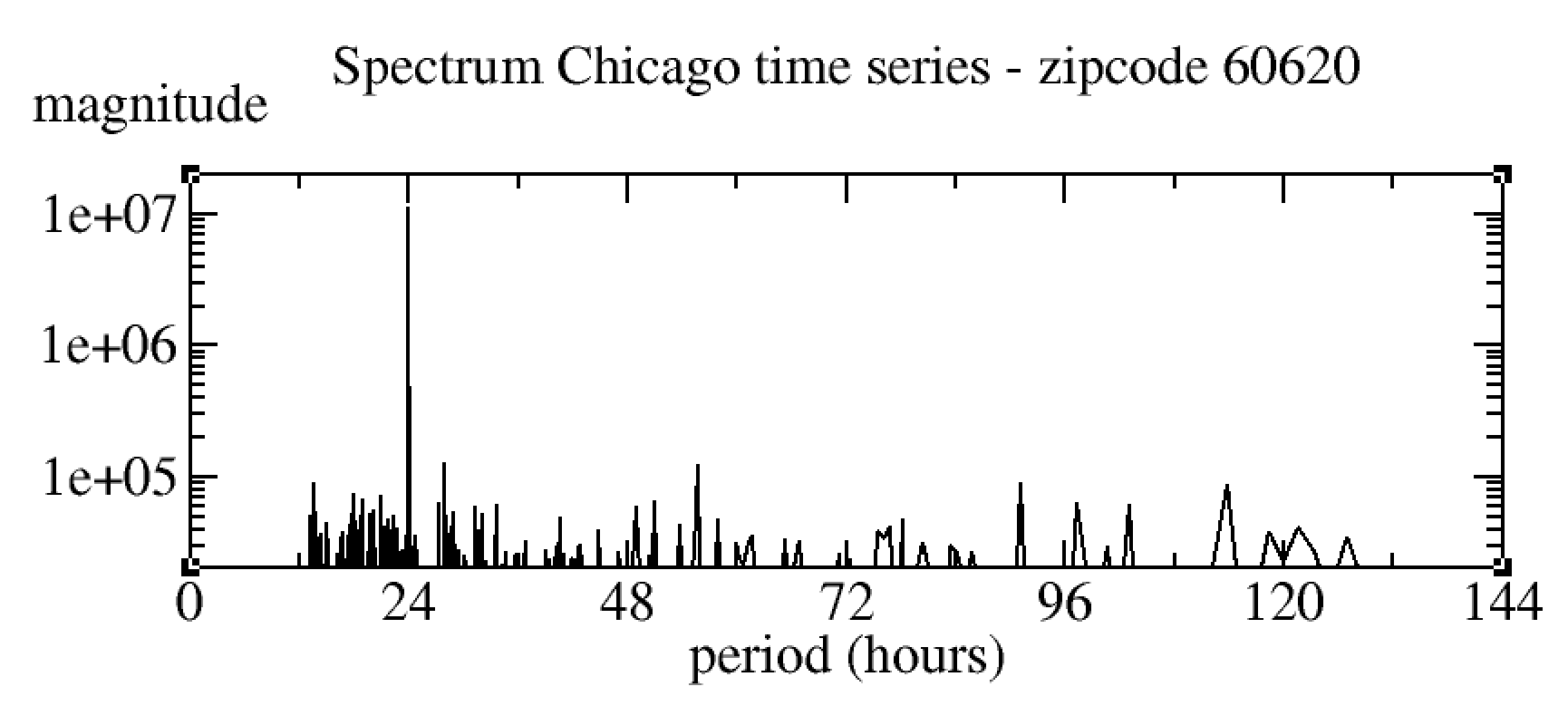}
\caption{Spectrum of the time series from Fig.~\ref{CrimeTimeSeries}.}
\label{spectrum}
\end{figure}

Figure \ref{CrimeTimeSeries} plots the hourly crime intensities over the entire CHI and a randomly selected zip code region. Though the time series are quite noisy, the spectrum exhibits clear diurnal periodicity as shown in Fig. \ref{spectrum}.

%However, as is shown in Fig. \ref{Spectrum-CrimeTimeSeries}, it is easy to see that the spectrums of all four time series are peaked at 24, which indicates these time %series all have periodic signal with period equal to one day. Additional tests show this is true for all regions in LA and Chicago. Furthermore, the spectrums of the time %series of the selected zipcode regions are noisier than that of the entire city.

%%%% Proofread by Xiyang from beginning up to here...
%\begin{figure}
%\centering
%\begin{tabular}{cccc}
%\hskip -0.65cm
%\includegraphics[width=0.30\columnwidth]{Chi_Whole_Jan28.png}&
%\hskip -0.55cm
%\includegraphics[width=0.30\columnwidth]{Chi_Zipcode_Jan28.png}&%\\
%\hskip -0.55cm
%%(a)&(b)\\
%\includegraphics[width=0.30\columnwidth]{Chi_Whole_Spectrum_Jan28.png}&
%\hskip -0.55cm
%\includegraphics[width=0.30\columnwidth]{Chi_Zipcode_Spectrum_Jan28.png}\\
%(a)&(b)&(c)&(d)\\
%\end{tabular}
%\caption{Example plot of crime time series and corresponding spectrum. (a) and (b) are the hourly crime intensities of the entire CHI and the zipcode region with postal code 60620 over the year 2015, respectively; (c) and (d) are the spectrum of the corresponding time series.}
%\label{CrimeTimeSeries}
%\end{figure}

% \begin{figure}
% \centering
% \begin{tabular}{cc}
% \includegraphics[width=0.40\columnwidth]{Chi_Whole.png}&
% \includegraphics[width=0.40\columnwidth]{Chi_Zipcode.png}\\
% (a)&(b)\\
% \includegraphics[width=0.40\columnwidth]{LA_Whole.png}&
% \includegraphics[width=0.40\columnwidth]{LA_Zipcode.png}\\
% (c)&(d)\\
% \end{tabular}
% \caption{The plot of crime time series. (a) and (b) are the hourly crime intensities of the whole CHI and the zipcode region with postal code 60620 over the year 2015, %respectively; (c) and (d) are the hourly crime intensities of the whole LA and zipcode region with postal code 90003 over the years 2014 and 2015, respectively.}
% \label{CrimeTimeSeries}
% \end{figure}
%
% \begin{figure}
% \centering
% \begin{tabular}{cc}
% \includegraphics[width=0.40\columnwidth]{Chi_Whole_Spectrum.png}&
% \includegraphics[width=0.40\columnwidth]{Chi_Zipcode_Spectrum.png}\\
% (a)&(b)\\
% \includegraphics[width=0.40\columnwidth]{LA_Whole_Spectrum.png}&
% \includegraphics[width=0.40\columnwidth]{LA_Zipcode_Spectrum.png}\\
% (c)&(d)\\
% \end{tabular}
% \caption{Spectrums of the crime time series corresponding to the ones depicted in Fig.\ref{CrimeTimeSeries}.}
% %(a) and (b) are spectrum of the hourly crime intensities over the whole CHI and the zip code region with postal code 60620 over the year 2015, respectively; (c) and (d) % depict the spectrum of the hourly crime intensities over the whole LA and zip code region with zip code 90003 over the years 2014 to 2015, respectively.}
% \label{Spectrum-CrimeTimeSeries}
% \end{figure}

\subsubsection{Statistical Analysis of Crime Data}
% Hawkes process simulation and deficiency in forecasting.
Evidence suggests that crime is self-exciting \cite{RN3516}, which is reflected in the fact that crime events are clustered in time. The arrival of crimes can be modeled as a Hawkes process (HP) \cite{Mohler:2011JASA} with a general form of conditional intensity function:
\begin{equation}
\label{Hawkes-1D}
\lambda(t) = \mu + a\sum_{t_i\leq t} g(t-t_i)
\end{equation}
where $\lambda(t)$ is the intensity of events arrival at time $t$, $\mu$ is the endogenous or background intensity, which is simply modeled by a constant, $a$ is the self-exciting rate, and $g(t)$ is a kernel triggering function. In \cite{Zipkin:2016}, it found that an exponential kernel, i.e., $g(t)=w\exp(-wt)$ where $\frac{1}{w}$ models the average duration of the influence of an event, is a good description of crime self-excitation. To calibrate the HP, we use the expectation-maximisation (EM) algorithm \cite{Veen:2008}. Simulation of the HP is done via a simple thinning algorithm \cite{Laub:2015}.

% Plot one selected data and simulated and averaging over multiple paths
% The stochastic process that fit the data and explain it.

%The HP fits to the crime intensity time series to the zipcode regions 90003 and 60620 with exponential kernel are given, respectively, as:
%$$
%\lambda(t) = 0.4982 + \sum_{t_i<t}0.2976*25.8143*\exp(-25.8143*(t-t_i))
%$$
%and
%$$
%\lambda(t) = 0.7562 + \sum_{t_i<t}0.4673*31.6301*\exp(-31.6301*(t-t_i))
%$$
%It is shown that on average, exogenously there will be 0.4982 and 0.7562 offspring crimes happen in the region 90003 and 60620, respectively. The crimes in the region %60620 is more self-exciting that 90003. The duration of the influences of the given events to its offsprings are roughly the same for both regions.

The HP fits to the crime time series in zip code region 60620 yields
%$$
%\lambda(t) = 0.4982 + \sum_{t_i<t}0.2976*25.8143*\exp(-25.8143*(t-t_i))
%$$
%and
$\lambda(t) = 0.7562 + \sum_{t_i<t}0.4673*31.6301*\exp(-31.6301*(t-t_i))$,
which shows that on average, each crime will have 0.4673 offspring. Furthermore, we noticed that the duration of the influence is roughly a constant over different zip code regions.

%The crimes in the region 60620 is more self-exciting that 90003. The duration of the influences of the given events to its offsprings are roughly the same for both regions.

\begin{figure}
%\centering
\begin{tabular}{ccc}
%\includegraphics[width=0.40\columnwidth]{figs/Nov2015_FirstTwoWeeks_60620.png}&
%\includegraphics[width=0.40\columnwidth]{figs/Nov2015_FirstTwoWeeks_90003.png}\\
%(a)&(b)\\
%\includegraphics[width=0.40\columnwidth]{figs/Nov2015_FirstTwoWeeks_60620_Hawkes.png}&
%\includegraphics[width=0.40\columnwidth]{figs/Nov2015_FirstTwoWeeks_90003_Hawkes.png}\\
%(c)&(d)\\
%\includegraphics[width=0.40\columnwidth]{figs/Nov2015_FirstTwoWeeks_60620_Hawkes_Average.png}&
%\includegraphics[width=0.40\columnwidth]{figs/Nov2015_FirstTwoWeeks_90003_Hawkes_Average.png}\\
%(e)&(f)\\
\includegraphics[width=0.33\columnwidth]{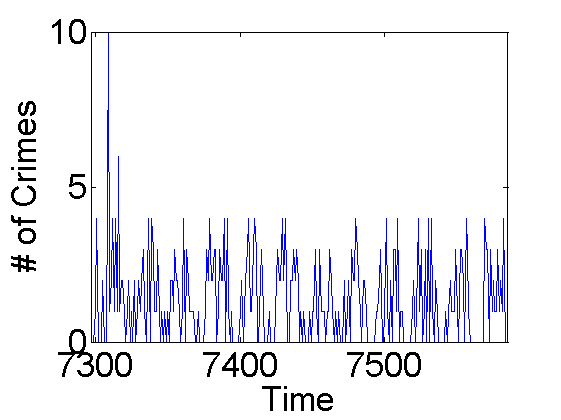}&
\hskip -0.5cm
\includegraphics[width=0.33\columnwidth]{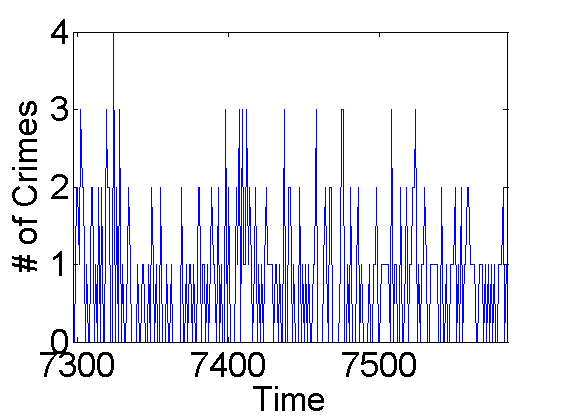}&
\hskip -0.5cm
\includegraphics[width=0.33\columnwidth]{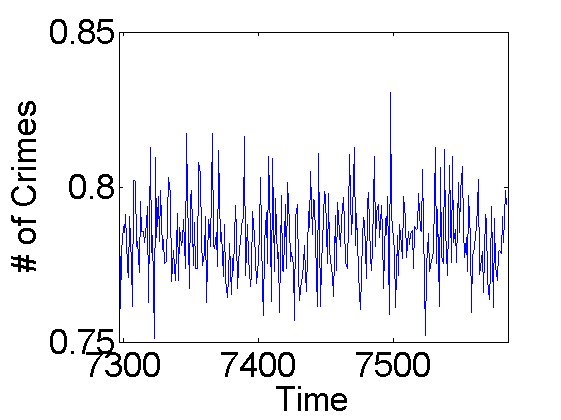}\\
(a) Exact&(b) one path sampled&(c) ensemble 5000\\
%\includegraphics[width=0.30\columnwidth]{Nov2015_FirstTwoWeeks_90003.png}&
%\includegraphics[width=0.30\columnwidth]{Nov2015_FirstTwoWeeks_90003_Hawkes.png}&
%\includegraphics[width=0.30\columnwidth]{Nov2015_FirstTwoWeeks_90003_Hawkes_Average.png}\\
%(d)&(e)&(f)\\
\end{tabular}
\caption{Exact and simulated hourly crime intensities for CHI 60620 in the first two weeks of Nov 2015. (a), (b), and (c) depict the exact, one path sampled from the HP, and the ensemble average of 5000 paths, respectively.
%In these plots, we regard the first time slots of both cities as 1.
}
\label{Hawkes-Simulation-TimeSeries}
\end{figure}

Figure \ref{Hawkes-Simulation-TimeSeries} shows the exact and simulated crime intensities in the first two weeks in Nov 2015 over zip code region 60620. Both the exact and simulated time series demonstrate clustered behavior, which confirms the assumption that crime time series is self-exciting, and supports the contention that the HP is a suitable model. However, the simulate intensity peaks are shifted relative to the exact ones. If we use the HP to do the crime forecasting, we typically do an ensemble average of many independent realizations of the HP, as is shown in panel (c). However, this ensemble average differs hugely from the exact crime time series. To capture fine scale patterns we will use DNN.

%Overall, HP provides a good way to analyze the macro-scale statistical properties of the crime events, it is inappropriate for real time crime forecasting. To remedy this % deficiency, we will use a DNN to approximate the micro-scale features of the crime time series.

%\begin{remark}
%To sample the paths from HP, we take an "on-the-fly" approach. During the simulation, we put all the simulated events to the historical filtration as ancestors of the %coming events.
%\end{remark}

\subsection{Traffic Data}
We also study the ST distribution of traffic data \cite{Junbo:2017}. The data contains two parts: taxi records from Beijing (TaxiBJ) and bicycle data from New York city (BikeNYC). Basic analyse in \cite{Junbo:2017} shows periodicity, meteorological dependence, and other basic properties of these two datasets. The time span for TaxiBJ and BikeNYC are selected time slots from 7/1/2013 to 4/10/2016 and the entire span 4/1/2014-9/30/2014, respectively. The time intervals are 30 minutes and one hour, respectively. Both data are represented in Eulerian representations with lattice sizes to be 32$\times$32 and 16$\times$8, respectively. Traffic flow prediction using this traffic dataset will be selected as benchmark to evaluate our model.

\section{Algorithms and Models}
\label{Algorithms}
Our model contains two components. The first part is a graph representation for the spatio-temporal evolution of the data, where the nodes of the graph are selected to contain sufficient predictable signals, and the topological structure of the graph is inferred from self-exciting point process model. The second component is a DNN to approximate the temporal evolution of the data, which has good generalizability. The advantages of a graph representation are two-fold: on the one hand, it captures the irregularity of the spatial domain; on the other hand, it can capture versatile spatial partitioning which enables forecasting  at different spatial scales. In this section, we will present the algorithms for modeling and forecasting the ST sparse unstructured data. The overall pipeline includes: STWG inference, data augmentation, and the structure and algorithm to train the DNN.

\subsection{STWG Representation for the ST Data}
The entire city is partitioned into small pieces with each piece representing one zip code region, or other small region. This partitioning retains geographical cohesion. In the STWG, we associate each geographic region with one node of the graph. The inference of the graph topological structure is done by solving the maximal likelihood problem of the MHP. We model the time series on the graph by the following MHP $\{N_t^u| u =1, 2, \cdots, U\}$ with conditional intensity functions:
\begin{equation}
\label{Hawkes-MV}
\lambda_u(t) = \mu_u + \sum_{i: t_i<t}a_{uu_i}g(t-t_i),
\end{equation}
where $\mu_u \geq 0$ is the background intensity of the process for the $u$-th node and $t_i$ is the time at which the event occurred on node $u_i$ prior to time $t$. The kernel is exponential, i.e., $g(t) = w*\exp(-w*t)$.

We calibrate the model in Eq.(\ref{Hawkes-MV}) using historical data. Let $\boldmath{\mu}=(\mu_u|u=1, 2,\cdots, U)$ and $A=(a_{uu'}|u, u'=1, 2, \cdots, U)$. Suppose we have $m$ i.i.d samples $\{c_1, c_2, \cdots, c_m\}$ from the MHP; each is a sequence of events observed during a time period $[0, T_c]$. The events form a set of pairs $(t_i^c, u_i^c)$ denoting the time $t_i^c$ and the node $u_i^c$-th for each event.
The log-likelihood of the model is:
%The log-likelihood of the model parameter $\{\boldmath{\mu}, A\}$ is:
\begin{equation}
\label{Likelihood}
\mathcal{L}(\mu, A)=\sum_c\left(\sum_{i=1}^{n_c}\log \lambda_{u_i}(t_i^c)-\sum_{u=1}^U\int_0^{T_c}\lambda_u(t)dt\right).
\end{equation}

%\vskip -1.5cm
%\begin{strip}
%\begin{eqnarray}
%\label{Likelihood}
%\mathcal{L}(\boldmath{\mu}, A)=\sum_c\left(\sum_{i=1}^{n_c}\log \lambda_{u_i^c}(t_i^c)-\sum_{u=1}^U\int_0^{T_c}\lambda_u(t)dt\right)%\\ \nonumber
%=\sum_c\left[\sum_{i=1}^{n_c}\log\left(\mu_{u_i^c}+\sum_{t_j^c<t}a_{u_i^cu_j^c}g(t_i^c-t_j^c)\right) -T_c\sum_{u=1}^U \mu_n %-\sum_{u=1}^U\sum_{j=1}^{n_c}\int_0^{T_c-t_j}a_{uu_j^c}g(s)ds \right]
%\end{eqnarray}
%\end{strip}
%\vskip -0.01cm

Similar to the work by Zhou {\em et al} \cite{Zha:2013}, to ensure the graph is sparsely connected, we add an $L_1$ penalty, $\lambda |A|_1 = \lambda \sum_{uu'}|a_{uu'}|$, to the log-likelihood $\mathcal{L}$ in Eq.(\ref{Likelihood}):
%\begin{equation}
$\mathcal{L}_{\lambda}(\mu, A) = -\mathcal{L}+\lambda|A|_1$.
%\end{equation}
To infer the graph structure, we solve the optimization problem:
\begin{equation}
\label{OptimizationProblem}
{\rm argmin}_{\boldmath{\mu}, A} \mathcal{L}_{\lambda}(\boldmath{\mu}, A),\\ \nonumber
%&\mbox{s.t.}\ \ \\ \nonumber
\quad
\mbox{s.t.}\  \boldmath{\mu} \geq 0, \ {\rm and}\ \ A \geq 0,
\end{equation}
where $\boldmath{\mu} \geq 0$ and $A \geq 0$, both are defined element-wise. We solve the above constraint optimization problem by the EM algorithm. The $L_1$ constraint is solved by a split-Bregman liked algorithm \cite{Goldstein:2009}. For a fixed parameter $w$, we iterate between the following two steps until convergence is reached:
\begin{itemize}
\item {\bf E-Step:} Compute the exogenous or endogenous probability:% of an events:
$$
p_{ii}^c=\frac{\mu_{u_i^c}^{(k)}}{\mu_{u_i^c}^{(k)}+\sum_{j=1}^{i-1}a_{u_iu_j}^{(k)}g(t_i^c-t_j^c)}, %\ \
$$

$$
  p_{ij}^c=\frac{a_{u_iu_j}^{(k)}g(t_i^c-t_j^c)}{\mu_{u_i^c}^{(k)}+\sum_{j=1}^{i-1}a_{u_iu_j}^{(k)}g(t_i^c-t_j^c)}
$$

%$$
%p_{ii}^c=\frac{\mu_{u_i^c}^{(k)}}{\mu_{u_i^c}^{(k)}+\sum_{j=1}^{i-1}a_{u_iu_j}^{(k)}g(t_i^c-t_j^c)}
%$$
%$$
%p_{ij}^c=\frac{a_{u_iu_j}^{(k)}g(t_i^c-t_j^c)}{\mu_{u_i^c}^{(k)}+\sum_{j=1}^{i-1}a_{u_iu_j}^{(k)}g(t_i^c-t_j^c)}
%$$

\item {\bf M-Step:} Update parameters:
$$
\mu_u^{(k+1)}=\frac{1}{\sum_cT_c}\left(\sum_c\sum_{i=1, u_i^c=u}^{n_c}p_{ii}^c\right)
$$
$$
a_{uu'}^{(k+1)}=\left(\frac{a_{uu'}^{(k)}\sum_c\sum_{i:u_i^c=u}\sum_{j<i, u_j^c=u'}p_{ij}^c}{\sum_c\sum_{j: u_j^c=u'}\int_0^{T_c-t_j^c} g(t)dt} \right)^{1/2}
$$
$$
a_{uu'}^{(k+1)}={\rm shrink}_\lambda(a_{uu'}^{(k+1)}).
$$
\end{itemize}
The above EM algorithm is of quadratic scaling, which is infeasible for our datasets. To reduce the algorithm's complexity, instead of considering all events before a given time slot, we do a simple truncation in the E-step based on the localization of the exponential kernel.
This truncation simplifies the algorithm from quadratic scaling to almost linear scaling.
%Overall, we formulate the algorithm in the Algorithm \ref{GraphInference}.
%In the inference of the STWG for both CHI and LA, we set the hyper-parameter $\lambda$ to be $0.01$.
In the inference of the STWG, we set the hyper-parameter $\lambda$ to be $0.01$.
%\begin{algorithm}
%\caption{Algorithm for the Latent Graph Inference.}\label{GraphInference}
%\begin{algorithmic}
%    \State \textbf{Input:} The historical events pairs $\{(t_i^c, u_i^c)\}$, the hyper-parameter $\lambda$.
%    \State \textbf{Output:} The inferred parameters $\boldmath{\mu}$ and $A$, and the log-likelihood $\mathcal{L}$.
%\For {$w$ in the set of grid search}
%    \While {not converge}
%    \State {\bf E-Step:}
%
%Compute the probability of an events to be exogenous or endogenous:
%$$
%p_{ii}^c=\frac{\mu_{u_i^c}^{(k)}}{\mu_{u_i^c}^{(k)}+\sum_{j=1}^{i-1}a_{u_iu_j}^{(k)}g(t_i^c-t_j^c)}
%$$
%$$
%p_{ij}^c=\frac{a_{u_iu_j}^{(k)}g(t_i^c-t_j^c)}{\mu_{u_i^c}^{(k)}+\sum_{j=1}^{i-1}a_{u_iu_j}^{(k)}g(t_i^c-t_j^c)}
%$$
%
%	\State {\bf M-Step:}
%
%Update parameters:
%$$
%\mu_u^{(k+1)}=\frac{1}{\sum_cT_c}\left(\sum_c\sum_{i=1, u_i^c=u}^{n_c}p_{ii}^c\right)
%$$
%$$
%a_{uu'}^{(k+1)}=\left(\frac{a_{uu'}^{(k)}\sum_c\sum_{i:u_i^c=u}\sum_{j<i, u_j^c=u'}p_{ij}^c}{\sum_c\sum_{j: u_j^c=u'}\int_0^{T_c-t_j^c} g(t)dt} \right)^{1/2}
%$$
%$$
%a_{uu'}^{(k+1)}={\rm shrink}_\lambda(a_{uu'}^{(k+1)}).
%$$
%	\EndWhile
% 	\State Compute the log-likelihood $\mathcal{L}(w)$ given by Eq.(\ref{Likelihood}) for the given $w$.
%\EndFor
%\State Return $\boldmath{\mu}$ and $A$ solved by$
%{\rm argmax}_{w} \mathcal{L}(w).$
%\end{algorithmic}
%\end{algorithm}

\subsubsection{Results on STWG Inference}
\begin{figure}
\centering
\includegraphics[width=1.0\columnwidth]{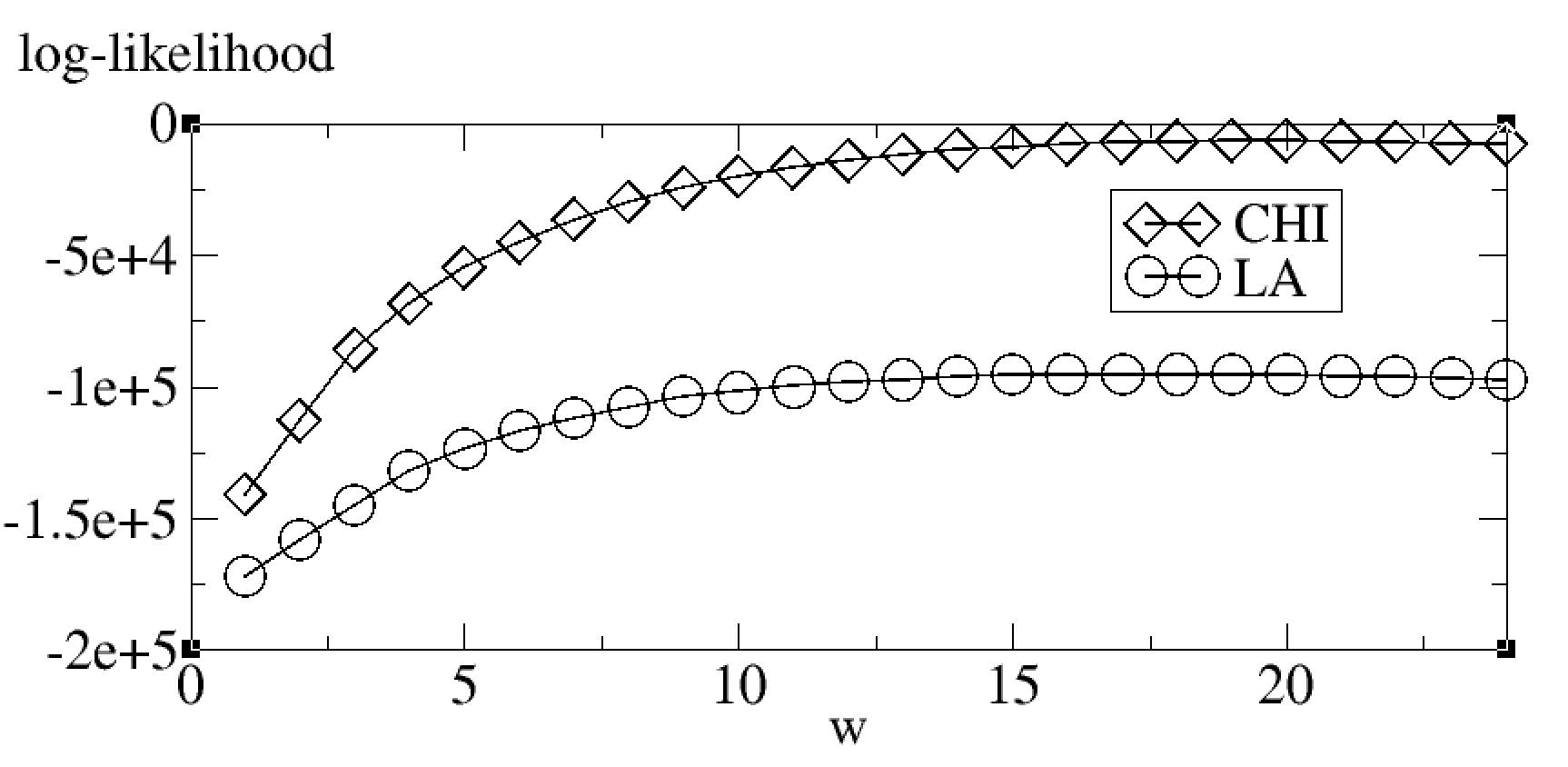}
\caption{Plot of $w$ vs log-likelihood. The maximum value occurs at $w=20$ and $w=18$ respectively for CHI and LA.}
\label{Log-likelihood}
\end{figure}

\begin{figure}
\centering
\begin{tabular}{cc}
\includegraphics[width=0.48\columnwidth]{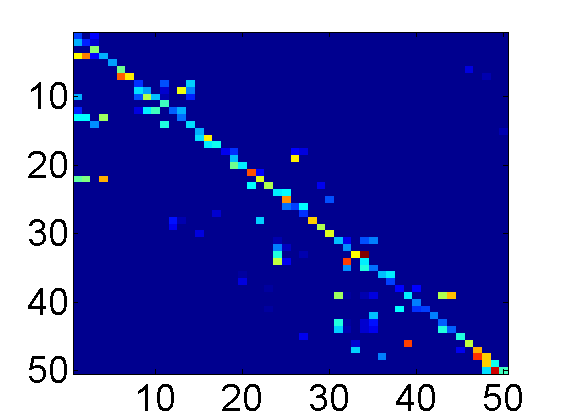}&
\includegraphics[width=0.48\columnwidth]{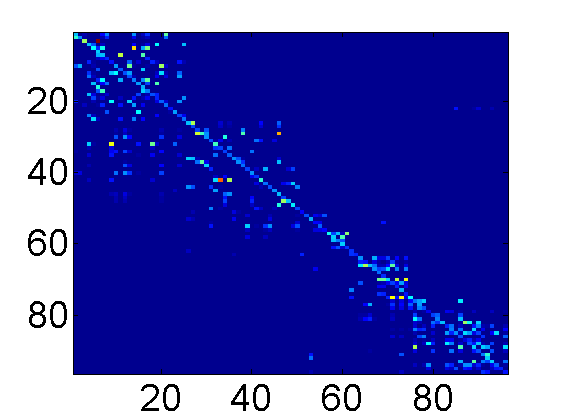}\\
(a) CHI&(b) LA\\
\end{tabular}
\caption{Image plot of the self and mutual excitation matrix $A$ for the cities CHI and LA.}
\label{Graph-Structure}
\end{figure}

% Insert the log-likelihood here.
Due to the high condition number of the log-likelihood function with respect to the parameter $w$ \cite{Zipkin:2016}, we perform a simple grid search to find the optimal $w$ (see Fig. \ref{Log-likelihood}). The likelihood functions are maximized when $w$ is 20 and 18 for CHI and LA, respectively. The similarity between the optimal duration parameters for Chicago and Los Angeles suggest that the duration of the self-excitation is an intrinsic property of crime. The optimal self-excitation parameters sets $A$ for two cities are plotted in Fig.\ref{Graph-Structure}. The diagonal in Fig. 6 reflects the intensity of self-excitation within a single node of the graph (i.e., zip code region). Off-diagonal entries reflects self-excitation of crime between nodes of the graph. Only nodes that demonstrate self-excitation above a threshold theta are connected by an edge in the final graph.
%A snapshot of the inferred STWG for LA is shown in Fig.\ref{Graph-LA-Snapshot}.}

%The inferred matrix $A$ is sparse which confirms the STWG
%is sparse connected. In general, the self-exciting dominates.
% the mutual-exciting.

% \begin{figure}
% \centering
% \begin{tabular}{c}
% \includegraphics[width=0.8\columnwidth]{Graph_LA_Snapshot.png}\\
% \end{tabular}
% \caption{A snap-shot of the STWG representation for the LA. The nodes represent the regions, the solid lines represent the strength of interaction from %source nodes to sink nodes, and the dashed line indicates the un-plotted interactions.}
% \label{Graph-LA-Snapshot}
% \end{figure}

% Discuss sparsity and self-mutual excitation.

\subsubsection{Effectiveness of STWG Inference}
% Xiyang: I will write this part
We validate the efficacy of the inference algorithm on a synthetic problem. To generate the synthetic data, a random graph $G$ is first generated with a fixed level of sparsity on a fixed set of nodes $i = 1, \dots, U$.
A MHP $E$ is then simulated for a fixed amount of time $T$ with randomly generated background rate $\mu_i$, and excitation rates $a_{i,j}$ supported
on the graph $G$. We use the aforementioned algorithm %maximum likelihood approach with an $L_1$ penalty
%in section \ref{sec:introduction}
to infer the coefficients $\hat{a}_{i,j}$. To obtain the underlying graph structure, there is an edge connected from node $j$ to $i$ if and only if $\text{Sign}(\hat{a}_{i,j} - \theta)>0$, where $\theta$ is a threshold that determines the sparsity of the inferred weighted graph.
To evaluate the efficacy of the inference algorithm, we vary the threshold $\theta$ to obtain a ROC curve, %\FB{(unresolved Jeff's comment: Is ROC obvious?)},
where a connection between two nodes $i$ and $j$ is treated as positive and vice versa. The area under the ROC curve (AUC) will be a metric on the performance of the algorithm.

For the experiments, we generate a directed and fully connected graph $G$ with $N=30$ nodes, and keep each edge $e_{ij}$ with probability $s = 0.1, 0.2, \dots, 0.5$, where $s$ denotes the sparsity level of the graph. We generate at random $\mu_i\sim Unif([0, 0.1])$ and $a_{i,j} \sim Unif([0.02, 0.1])$ for $i,j$ connected in $G$, and $0$ otherwise. And we check the stability condition in the spectral norm where $\rho(A)<1$. A HP is then simulated with $T = 3 \times 10^4$.
In crime networks, it is reasonable to assume that the interactions $a_{ij}$ are local, and hence we may start out with a reduced set of edges during the inference procedure to increase accuracy of the network recovery. Therefore, in addition to recovering the network structure from a fully connected graph, we also test the inference algorithm on a set of reduced edges that contain the ground truth. For simplicity, we randomly choose $200$ and $400$ edges from the graph and add them to the true network structure at initialization.  We observe that the inference algorithm is able to obtain an AUC of around 0.9 across all levels of sparsity, with large increases in performance if the graph prior is narrower.

% Xiyang: add imagesc plot of true and sparse graphs.
%\begin{table}[!t]
%\caption{AUC of the ROC curve for the graph inference problem. Rows denote the sparsity of the ground truth graph, columns are different prior knowledge for network structure for the inference algorithm. }
%\renewcommand{\arraystretch}{1.3}
%\centering
%\begin{tabular}{|c|c|c|c|c|c|c|}\hline
% Prior/Sparsity& 0.1& 0.2 & 0.3& 0.4& 0.5 \\ \hline
%Null&0.9001   &0.8969    &0.8840    &0.9145  & 0.9099 \\ \hline
%GT + 200&0.9891  &0.9865   &0.9824   &0.9806  & 0.9857      \\ \hline
%GT + 400&0.9693  &0.9556   &0.9620   & 0.9467 &  0.9542      \\ \hline
%\end{tabular}
%\end{table}

\begin{table}[!t]
\caption{AUC of the ROC curve for the graph inference problem. Rows denote the sparsity of the ground truth graph, columns are different prior knowledge for network structure for the inference algorithm. }
\renewcommand{\arraystretch}{1.3}
\centering
\begin{tabular}{|c|c|c|c|c|c|c|}\hline
 Prior/Sparsity& 0.1& 0.2 & 0.3& 0.4& 0.5 \\ \hline
Null&0.900   &0.897    &0.884    &0.915  & 0.910 \\ \hline
GT + 200&0.989  &0.987   &0.982   &0.981  & 0.986      \\ \hline
GT + 400&0.969  &0.956   &0.962   &0.947 & 0.954      \\ \hline
\end{tabular}
\end{table}

\begin{figure}
\centering
\begin{tabular}{ccc}
\hskip -0.7cm
\includegraphics[width=0.37\columnwidth]{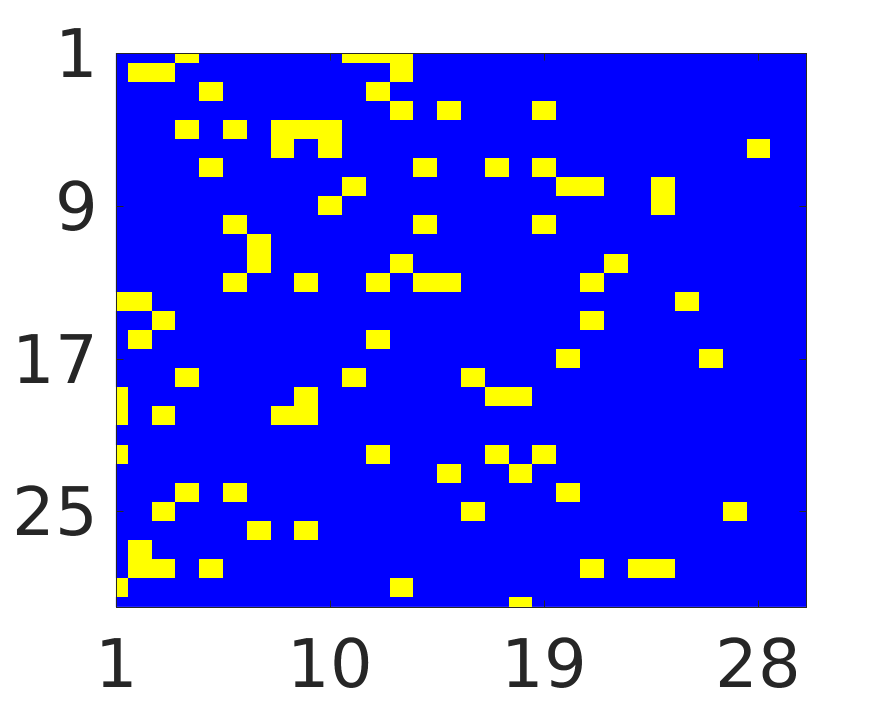}&
\hskip -0.4cm
\includegraphics[width=0.37\columnwidth]{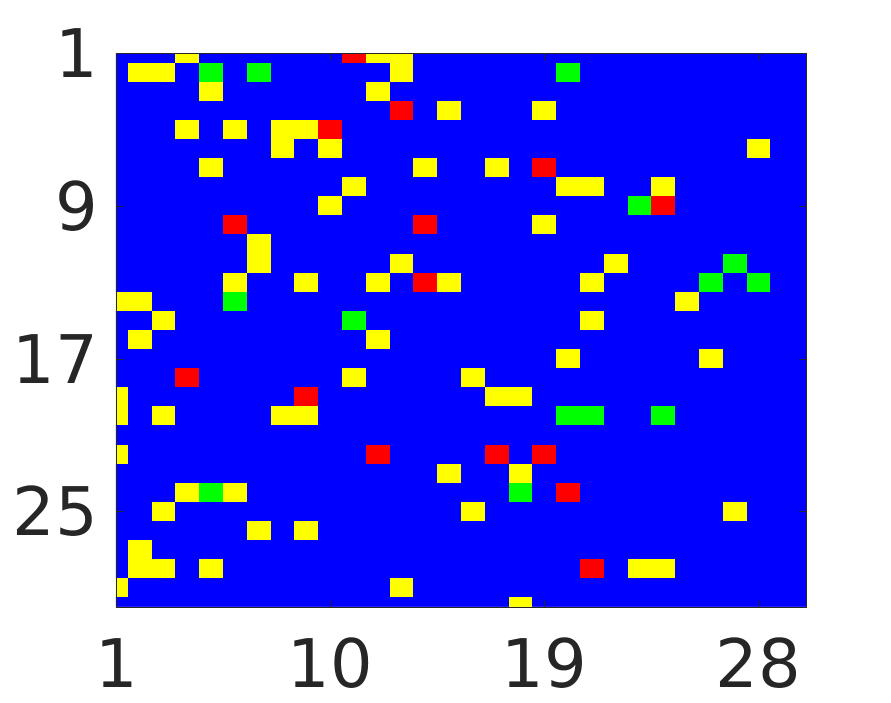}&
\hskip -0.4cm
\includegraphics[width=0.37\columnwidth]{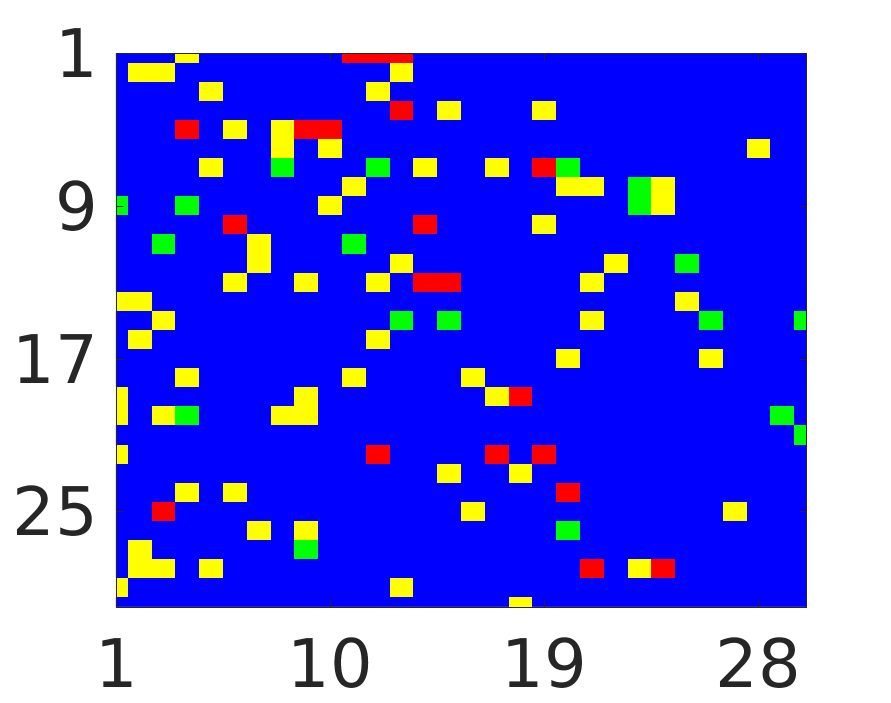}\\
(a) GT &(b) GT+200&(c) GT+400\\
\end{tabular}
\caption{Visualization of the ground truth and inferred graph for the synthetic data. The inferred graphs were obtained by thresholding $a_{ij}$ to match the sparsity level of the original network. The true positives, true negatives, false positives, false negatives are color coded in yellow, blue, red, green respectively. }
\end{figure}

\subsection{Data Augmentation - Single Node Study}
We consider data augmentation to boost the performance of the DNN for sparse data forecasting, with single zip code crime forecasting as an illustration.
In our previous work \cite{BaoWang:2017DL1,BaoWang:2017DL2}, when dealing with crime forecasting on a square grid, we noticed the DNN poorly approximates the crime intensity function. However, it does approximate well the diurnal cumulated crime intensity, which has better regularity. According to the universal approximation theorem \cite{Cybenko:1989}, the DNN can approximate any continuous function with arbitrary accuracy. However, the crime intensity time series is far from a continuous function due to its spatial and temporal sparsity and stochasticity. Mathematically, consider the diurnal time series $\{x(t)\}$ with period $T$. We map $\{x(t)\}$ to its diurnal cumulative function via the following periodic mapping:
\begin{equation}
\label{Map}
y(t)=\int_{nT}^t x(s) ds\doteq I(x(t)),
\end{equation}
for $t\in[nT, (n+1)T)$, this map is one-to-one.% which is easy to recover the original signal.

We also super-resolve the diurnal cumulated time series $\{y(t)\}$ to constract an augmented time series $\{\hat{y}(T)\}$ on half-hour increments via linear interpolation.
The new time series has a period of $\hat{T}=2T-1$. In the time interval $[n\hat{T}, (n+1)\hat{T})$ it is defined as:
%\begin{equation}
%\label{TS}
%\hat{y}(t) = \begin{cases}
%y(t) &\text{$t=nT+2k$}\\
%\frac{y(t)+y(t+1)}{2} &\text{$t=nT+2k+1$,}
%\end{cases}
%\end{equation}
\begin{equation}
\label{TS}
\hat{y}(t) = \begin{cases}
y(nT+k) &\text{$t=n\hat{T}+2k$}\\
\frac{1}{2}[y(nT+k)+y(nT+k+1) ]&\text{$t=n\hat{T}+2k+1$,}
\end{cases}
\end{equation}
for $k=0, 1, \cdots, T-1$.
It is worth noting the above linear interpolation is completely local. In the following DNN training procedure it will not lead to information leak.% with the closest prior be skipped.
%The aforementioned data augmentation scheme is illustrated in the Fig.\ref{Data-Augmentation}.
%
%\begin{figure}
%\centering
%\begin{tabular}{ccc}
%\includegraphics[width=0.3\columnwidth]{LA_Zipcode_TwoMonths.png}&
%\includegraphics[width=0.3\columnwidth]{LA_Zipcode_CDF_TwoMonths.png}&
%\includegraphics[width=0.3\columnwidth]{LA_Zipcode_CDFSupT_TwoMonths.png}\\
%(a)&(b)&(c)\\
%\end{tabular}
%\caption{Illustration of the data augmentation. Charts (a), (b), and (c) plot the crime intensity, diurnal cumulated intensity, and super resolved %diurnal cumulated intensity function of the crime time series on the zipcode region 90003 over the last two months of 2015, respectively.}
%\label{Data-Augmentation}
%\end{figure}

\subsubsection{Cascaded LSTM for Single Node Crime Modeling}
\begin{figure}% code for the flow chart
    \centering
    \begin{tikzpicture}[node distance=1cm]
    \node (input) [normal] {Input};
    %\node (lstm1) [normal, right of=input, xshift=1.0cm] {LSTM: Output (64,)};
    %\node (lstm2) [normal, right of=lstm1, xshift=1.0cm] {LSTM: Output (128,)};
    \node (lstm1) [normal, right of=input, xshift=0.8cm] {LSTM};
    \node (lstm2) [normal, right of=lstm1, xshift=0.8cm] {LSTM};
    \node (fc) [normal, right of=lstm2, xshift=0.8cm] {FC};
    \node (output) [normal, right of=fc, xshift=0.8cm] {Output};
    \draw [arrow] (input) -- (lstm1);
    \draw [arrow] (lstm1) -- (lstm2);
    \draw [arrow] (lstm2) -- (fc);
    \draw [arrow] (fc) -- (output);
    \end{tikzpicture}
    \caption{The architecture of the cascaded LSTM that used for single node data modeling.}
    \label{Stacked-LSTM}
\end{figure}
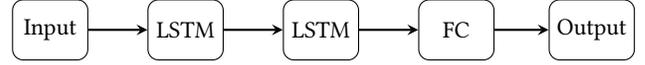

%In this part, we consider modeling the single node crime data by DNN.
The architecture of the DNN used to model single node crime is a simple cascaded LSTM as depicted in Fig.\ref{Stacked-LSTM}. The architecture contains two LSTM layers and one fully-connected (FC) layer, and represents the following function:
\begin{equation}
\label{DNN-Eq}
{\rm DNN}(x) = {\rm FC} \circ {\rm LSTM}_1 \circ {\rm LSTM}_2(x),
\end{equation}
where $x$ is the input. Generally, we can cascade $N$ layers of LSTM.%, which gives the approximator:
%begin{equation}
%\label{DNN-Eq2}
%\overline{{\rm DNN}}(x) = {\rm FC} \circ {\rm LSTM}_1 \circ {\rm LSTM}_{N-1} \circ \cdots \circ {\rm LSTM}_N(x).
%\end{equation}

%\begin{figure}
%\centering
%\begin{tabular}{c}
%\includegraphics[width=1.0\columnwidth]{LSTM.png}
%\end{tabular}
%\caption{The architecture of the cascaded LSTM that used for single node data modeling.}
%\label{Stacked-LSTM}
%\end{figure}

In the above cascaded architecture, all the LSTMs are equipped with 128 dimensional outputs except the first one with 64 dimensions. An FC layer maps the input tensor to the target value.
%Due to the DNN is applied to the super-resolved time series, to avoid data leaking, in both training and generalization we skip the value at the nearest previous time slot.
To avoid information leak when applying DNN to the super-resolved time series, we skip the value at the
nearest previous time slot in both training and generalization.

\begin{figure}
\centering
\begin{tabular}{c}
\includegraphics[width=0.85\columnwidth]{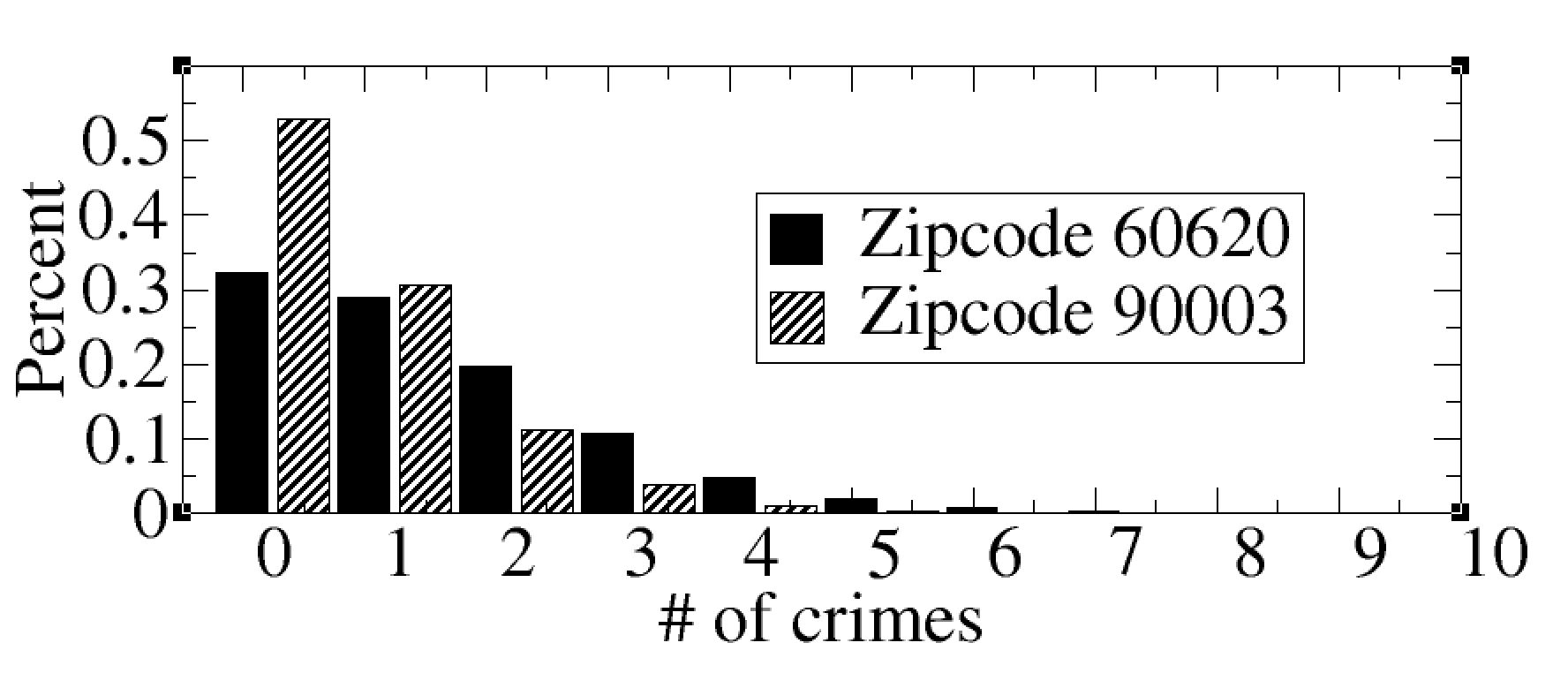}\\
\end{tabular}
\caption{Histogram of the hourly crime counts in zip code regions 60620, for the year 2015, and 90003, for 2014-2015.
%Top and bottom panels depict the hourly crime distribution of the year 2015 on the region 60620 and the years 2014-2015 on the region 90003.
}
\label{Histogram-Plot}
\end{figure}

Before fitting the historical crime intensities by the cascaded LSTMs, we first look at histograms of the crime intensities (Fig.\ref{Histogram-Plot}). The 99th percentiles of crime distributions are each less than six crimes.
%For the node 60620 the cases that with no more than 5 crimes takes more that 99 percent, similarly, for 90003 more than 99 percent time slots have less than 4 crimes.
%These histograms reveal the fact that considering
%the crime forecasting as a binary classification problem, where each
%time slot is classified as with crime or not, is not appropriate in that
%it fails to differentiate dangerous levels.  An efficient forecasting should reflects the histogram distributions.
This suggests that local crime intensity is important and one cannot use  a simple binary classifier.

%\begin{figure}
%\centering
%\begin{tabular}{cc}
%\includegraphics[width=0.4\columnwidth]{OneYear60620_Histogram.png}&
%\includegraphics[width=0.4\columnwidth]{TwoYears90003_Histogram.png}\\
%(a)&(b)\\
%\end{tabular}
%\caption{Histogram plots of the crime distributions. (a) and (b) depict the hourly crime distribution of the year 2015 on the region 60620 and the years 2014-2015 on the region 90003.}
%\label{Histogram-Plot}
%\end{figure}

%\begin{figure}
%\centering
%\begin{tabular}{c}
%\includegraphics[width=0.85\columnwidth]{OneYear60620_Histogram_Jan28.png}\\
%(a)\\
%\includegraphics[width=0.85\columnwidth]{TwoYears90003_Histogram_Jan28.png}\\
%(b)\\
%\end{tabular}
%\caption{Histogram of the hourly crime distributions in 60620 for the year 2015 (top) and 90003 for 2014-2015 (bottom).
%Top and bottom panels depict the hourly crime distribution of the year 2015 on the region 60620 and the years 2014-2015 on the region 90003.
%}
%\label{Histogram-Plot}
%\end{figure}

\begin{figure}
\centering
%\includegraphics[width=0.95\columnwidth]{Loss-v-Epochs.png}
%\vskip -1.1 in
\includegraphics[width=0.95\columnwidth, trim={0 0 0 4in},clip]{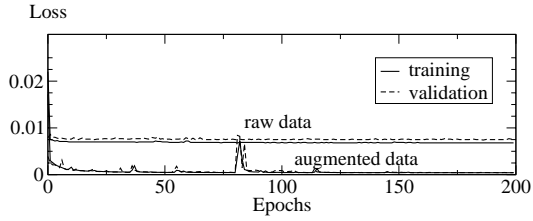}
\caption{Training procedures of the different scenarios on the node 90003.  Evolution of the training and validation loss on the raw data and on the augmented data.}
\label{Training-Val-Loss}
\end{figure}

%% 3. Present the single node LSTM structure, and name it as cascaded LSTM
We adopt the two layers of LSTMs cascade, which is demonstrated in Fig.\ref{Stacked-LSTM} to fit the single node crime time series. To train the DNNs for a single node, we run 200 epochs with the ADAM optimizer \cite{adam_15}, starting from the initial learning rate 0.01 with decay rate $1e-6$.
Fig. \ref{Training-Val-Loss} shows the decay of the loss function for the raw crime time series and cumulated super-resolved (CS) time series in panels (a) and (b), respectively.
%It is seen that if the DNN is applied to the raw crime time series, the model cannot be well trained, where the loss function flattened with a very large value after only a few epochs. However, the DNN approximates well to the regularized time series.
It can be seen from the figure that DNN performs much better on
the regularized time series than the raw one, i.e. the loss function
reaches a much lower equilibrium. %\FB{(Fig. 10 looks not clear in pdf)}
%after the regularity of the crime time series improved by our data augmentation techniques, the DNN approximates to the time series much better.
%There is still a big bump in the evolution of the loss function, we believe this dues to the application of ADAM optimizer, we can overcome this issue by using appropriate stochastic gradient descent methods.

%% 4. Present the figs of the training procedure for cum-supT and raw data (90003 only)
%% Train 200 epochs with ADAM optimizer, start from learning rate 0.01 with decay rate 1.e-6.
%\begin{figure}
%\centering
%\begin{tabular}{cc}
%\includegraphics[width=0.4\columnwidth]{Raw_Loss_90003.png}&
%\includegraphics[width=0.4\columnwidth]{CST_Loss_90003.png}\\
%(a)&(b)\\
%\end{tabular}
%\caption{Training procedures of the different scenarios. (a): evolution of the training and validation loss on the original data. (b): evolution of the training and validation loss on the augumented data.}
%\label{Training-Val-Loss}
%\end{figure}

To show the advantage of the generalization ability of the DNN, we compare it with a few other approaches, including autoregressive integrated moving average (ARIMA), %\cite{box1970distribution},
K-nearest neighbors (KNN), and historical average (HA). We use these models to fit the historical data and perform one step forecasting in the same manner as our previous work \cite{BaoWang:2017DL2}. The root mean squared error (RMSE) between the exact and predicted crime intensities and optimal parameters for the corresponding model are listed in Table \ref{RMSE-90003}. Under the RMSE measure, DNN, especially on the augmented data, yields higher accuracy. The small RMSE reflects the fact that DNN approximates the crime time series with good generalization ability.

%Similar setting is set in NOLTA journal

%% 5. Present the table of RMSE over the single node for different method over LA 90003
\begin{table}[!h]
\renewcommand{\arraystretch}{1.3}
\centering
\caption{RMSE between the exact and predicted crime intensities over the last two months of 2015, in the region with zip code 90003. DNN(CS) denotes DNN model applied to the augmented data.}
\label{RMSE-90003}
\centering
\begin{tabular}{cc}
\hline
\textbf{Methods} & \textbf{RMSE (number of crimes)}\\
\hline
DNN 				& 0.858 \\
DNN (CS)   		    & 0.491 \\
ARIMA(25, 0, 26)   	& 0.941 \\
KNN (k=1)   		& 1.193 \\
HA   				& 0.904 \\
\hline
\end{tabular}
\end{table}

%% 6. Introduce matrix measure
%\ALB{However, the simple RMSE measure is insufficient to measure error appropriately for sparse data. Do HA and ARIMA really work better than KNN for crime forecasting? HA ignores the day to day variation, while ARIMA simply predicts the number of crimes to be all zeros after flooring. KNN predicts more useful information than both ARIMA and HA for the crime time series.}
However, the simple RMSE measure is insufficient to measure error appropriately for sparse data. Do HA and ARIMA really work better than KNN for crime forecasting? HA ignores the day to day variation, while ARIMA simply predicts the number of crimes to be all zeros after flooring. KNN predicts more useful information than both ARIMA and HA for the crime time series. We propose the following measure, which we call a ``precision matrix" (do not confuse it with the one used in statistics) $B$ be defined as:
$$
B=\left(
    \begin{array}{ccc}
      \beta_{10} & \cdots & \beta_{1n} \\
      \vdots & \vdots & \vdots \\
      \beta_{m0} & \cdots & \beta_{mn} \\
    \end{array}
  \right)
$$
where $\beta_{ij}=\frac{N_{ij}}{N_{i}}$, where $N_i\doteq \#\{t| x_t\geq i\}$, and $N_{ij}\doteq \#\{t| x_t\geq i\  {\rm and}\  (x_t^p\geq i\  {\rm or}\  x_{t-1}^p\geq i \ {\rm or}\ \cdots \ {\rm or}\  x_{t-j+1}^p\geq i)\}$,
for $i=1, 2, \cdots n$; $j=0, 1, \cdots, m$. Here $x_t$ and $x_t^p$ are the exact and predicted number of crimes at time $t$. This means for a
given threshold number of crimes $i$, we count the number of time intervals in the testing set at which the predicted and exact number of crimes both exceed this
threshold $i$, with an allowable delay $j$, i.e., the prediction is allowed within $j$ hours earlier than the exact
time.
%\ALB{time}.
%where $a_{ij}=\frac{\#\{t: x^p_t\geq i\}}{\#\{t: x_t\geq i\{}$, $i=1, 2, \cdots n$, $j=0, 1, \cdots, m$, represents the accuracy of the $i$ level anomaly prediction within $j$ hours delay.
%Which means that for a given threshold $i$, we count the number of time slots in the testing set at which the predicted and exact number of crimes are both more than $i$, the delay level $j$ means that the prediction is allowed within $j$ hours earlier than the exact.

%\ALB{This measure provides much better guidance for crime patrol strategies. For instance, if we forecast more crime to occur in a given patrol area, then we can assign more police resources to that area. This metric allows for a few hours of delay but penalizes against crimes happening earlier than predicted, due to the time irreversibility of forecasting. For the crime time series in nodes 60620 and 90003, we select $m=3$, $n=2$ and $m=5$, $n=4$, respectively. Fig.\ref{precison-matrix-90003} shows the precision matrices of the crime prediction by different methods, confirming that DNN together with data augmentation gives accurate crime forecasting. Meanwhile, the KNN also gives better results compared to other methods except the DNN with data augmentation. This corrects potential inaccuracies in the RMSE measure and confirms the spatial correlation of the crime time series.}

\begin{figure*}
\centering
\begin{tabular}{ccccc}
\includegraphics[width=0.4\columnwidth]{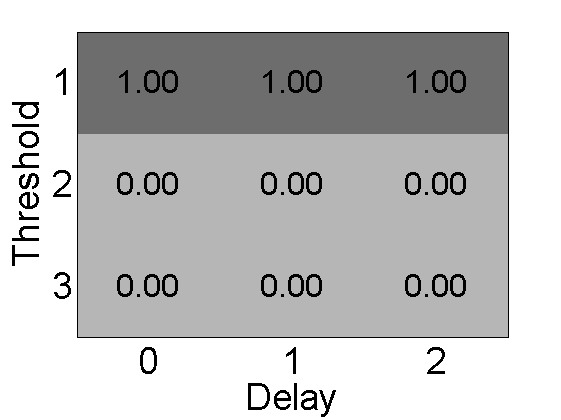}&
\hskip -0.5cm
\includegraphics[width=0.4\columnwidth]{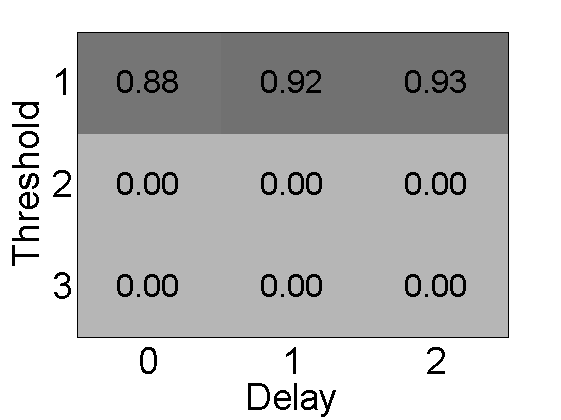}&
\hskip -0.5cm
\includegraphics[width=0.4\columnwidth]{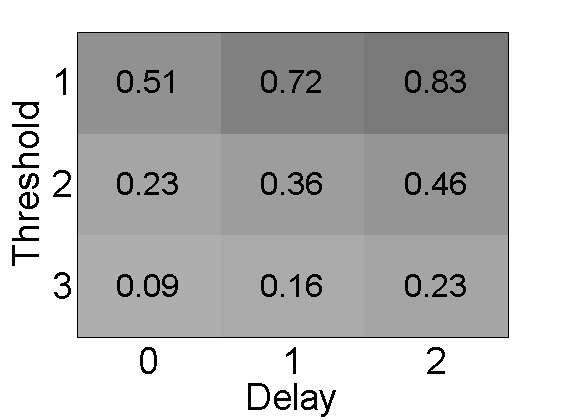}&
\hskip -0.5cm
\includegraphics[width=0.4\columnwidth]{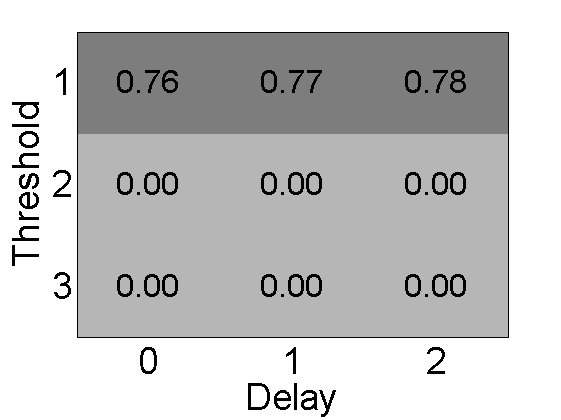}&
\hskip -0.5cm
\includegraphics[width=0.4\columnwidth]{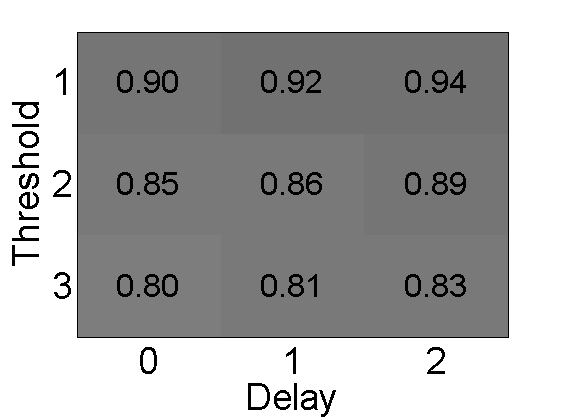}\\
(a) ARIMA&(b) HA&(c) KNN&(d) DNN original data &(e) DNN data augmentation\\
\end{tabular}
%\caption{\ALB{Precision matrix of the different predictors's performance in forecasting crime in 90003.}}
\caption{Precision matrix of the different predictors's performance in forecasting crime in 90003.}
\label{precison-matrix-90003}
\end{figure*}

This measure provides much better guidance for crime patrol strategies. For instance, if we forecast more crime to occur in a given patrol area, then we can assign more police resources to that area. This metric allows for a few hours of delay but penalizes against crimes happening earlier than predicted, due to the time irreversibility of forecasting. For the crime time series in nodes 60620 and 90003, we select $m=3$, $n=2$ and $m=5$, $n=4$, respectively. Fig.~\ref{precison-matrix-90003} shows the precision matrices of the crime prediction by different methods, confirming that DNN together with data augmentation gives accurate crime forecasting. Meanwhile, the KNN also gives better results compared to other methods except the DNN with data augmentation. This corrects potential inaccuracies in the RMSE measure and confirms the spatial correlation of the crime time series.

\begin{remark}
The precision matrix $B$ still has an issue in the case of over-prediction. Namely, this measure fails to penalize cases where the prediction is higher than the ground truth. However in those cases, the RMSE would typically be very large. Therefore, to determine if the sparse data is well predicted or not, we should examine both metrics.
\end{remark}

%% 7. Plot matrix measure of diferent menthod for LA 90003
%\begin{figure*}
%\centering
%\begin{tabular}{ccccc}
%\includegraphics[width=0.3\columnwidth]{Matrix_ARIMA.png}&
%\includegraphics[width=0.3\columnwidth]{Matrix_HA.png}&
%\includegraphics[width=0.3\columnwidth]{Matrix_KNN.png}&
%\includegraphics[width=0.3\columnwidth]{Raw_DNN_Matrix.png}&
%\includegraphics[width=0.3\columnwidth]{CST_DNN_Matrix.png}\\
%(a)&(b)&(c)&(d)&(e)\\
%\end{tabular}
%\caption{Precision matrix of the different predictors's performance in predicting the crimes in 90003 over the last two months of 2015. Panels (a)-(e) plot for the ARIMA, HA, KNN, DNN on the original data, and DNN with data augmentation, respectively.}
%\label{precison-matrix-90003}
%\end{figure*}

%% 8. Plot the exact and predicted time series, both CST and recovered on 90003

\begin{figure}
\centering
\begin{tabular}{cc}
\includegraphics[width=0.48\columnwidth]{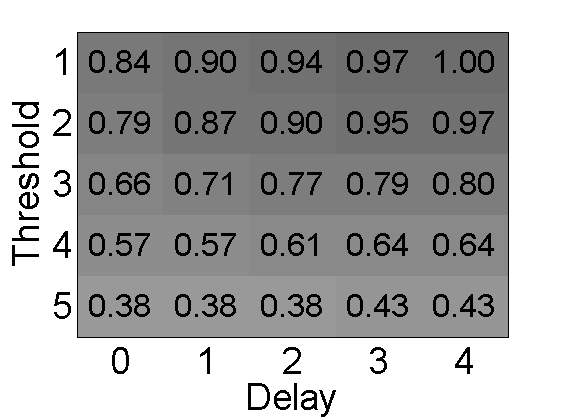}&
\hskip -0.5cm
\includegraphics[width=0.48\columnwidth]{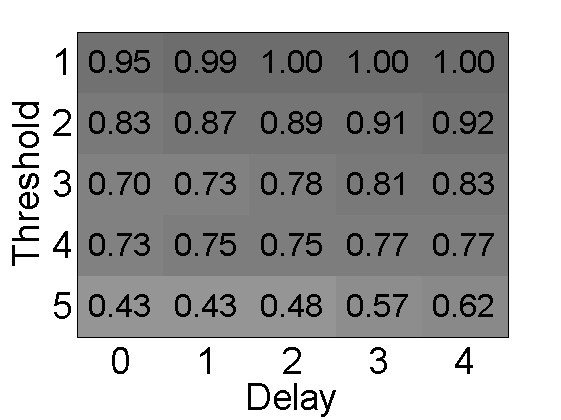}\\
(a) 2-layer LSTM (RMSE 0.901) &(b) 3-layer STM (RMSE 0.648)\\
\end{tabular}
%\caption{\ALB{Precision matrix of the cascaded two and three layers LSTMs for crime zip code 60620, with RMSE. }}
\caption{Precision matrix of the cascaded two and three layers LSTMs for crime zip code 60620, with RMSE.}
\label{precision-matrix-60620}
\end{figure}

Another merit of the DNN is that with sufficient training data, as the network goes deeper, better generalization accuracy can be achieved. To validate this, we test the 2 and 3 layers LSTM cascades on the node 60620 (see Fig.~\ref{precision-matrix-60620}).

%% 9. Chicago node, discuss when the network become deeper, result get better
%% Insert results here.

%%%%%%%%%%%%%%%%%%%%%%%%%%%%%%%%%%%%%%%%%%%%%%%%%%%%%%%%%%%%%%%%%%%%%%%%%%%%%%%%%%%%%%%%%%

\subsection{GSRNN for ST Forecasting}

\begin{figure}
\begin{tabular}{cc}
\hskip -1.5cm
\includegraphics[width=0.65\columnwidth]{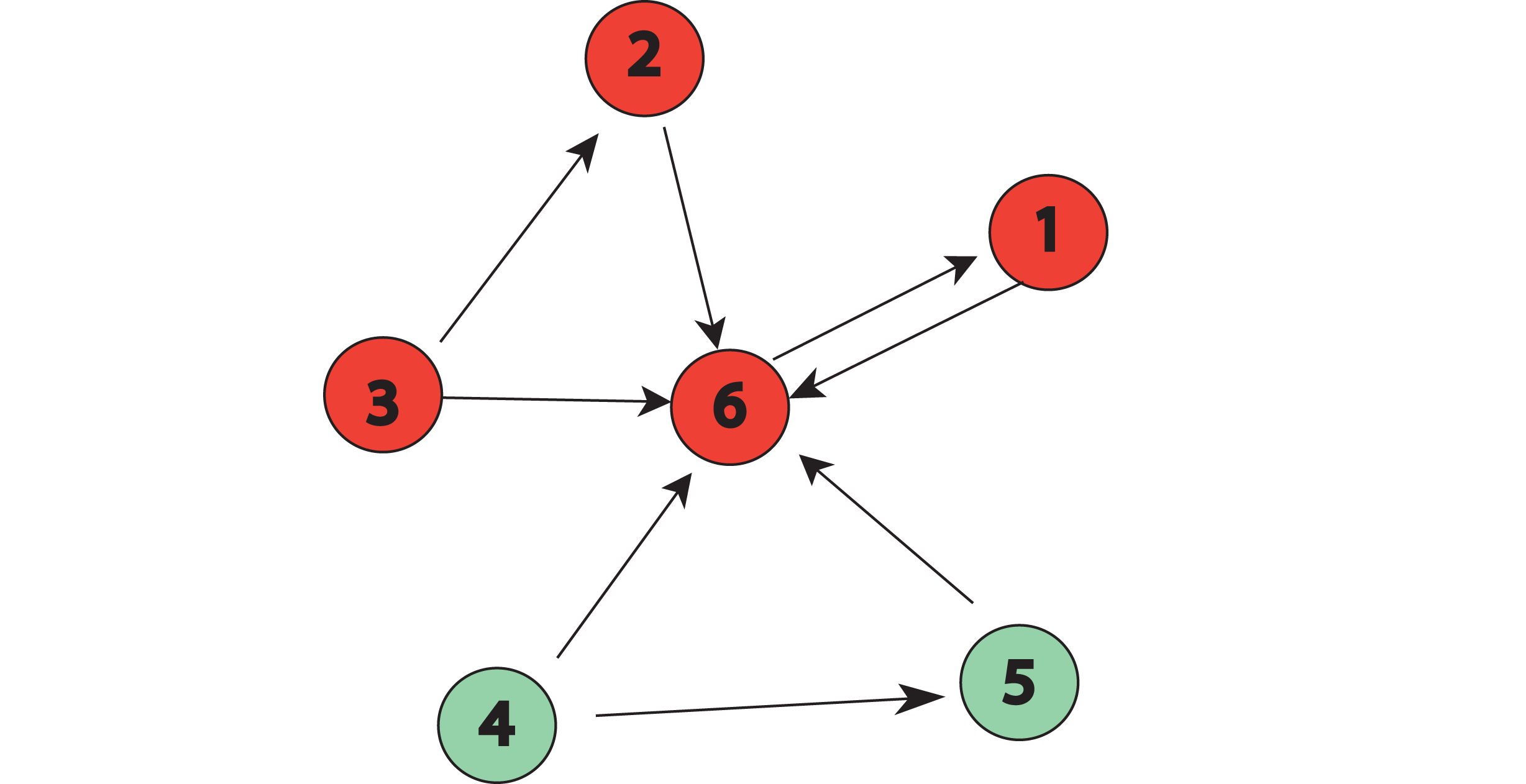}&
\hskip -1.5cm
\includegraphics[width=0.65\columnwidth]{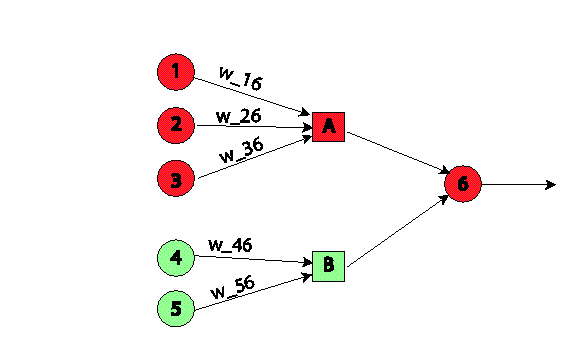}\\
(a)&(b)\\
\end{tabular}
\caption{Figure a) shows an example STWG inferred via the HP, where each color (red and green) denotes the class to which the node belongs. Figure b) depicts the feed-forward structure of our RNN network on a single node (node No.6).} %Each node that has a connection \emph{towards} node 6 first has its features combined via a weighted sum where the weights are the normalized mutual excitation rates from the HP. The combined features, are fed to an edge RNN that is then concatenated and fed into a node RNN.}
\label{fig:srnn}
\end{figure}
Our implementation of the GSRNN is based on the SRNN implementation in \cite{Jain:2016} (Fig.\ref{fig:srnn}), but differs in these key aspects: 1) We generalize the SRNN model to a directed graph, which is more suited to the ST data forecasting problem. 2) We use a weighted sum pooling based on the self-exciting weights from the MHP inference. 3) Due to the large number of nodes in the graph, we subsample each class of nodes for scalability.

To be more specific, suppose $i = 1, 2, \dots N$ are the nodes of the graph, and $X_i(t)$ denotes value of the time series at time $t$ for node $i$. We first deploy the STWG inference procedure to obtain the weighted directed graph, with weight on the edge connecting node $i$ and $j$ denoted as $w_{ij}$. With the same setup as in \cite{Jain:2016}, we partition the graph nodes to $K$ groups according to some criterion. We construct an ``input RNN'' $E^1_k$ for each class $k$, and an ``edge RNN'' $E^2_{k, l}$ for each class pair $(k,l)$ if $k\neq l$. For the forward pass, if node $i$ belongs to class $k$, we feed a set of historical data $\{X_{i}(t - p) | p = n_1, n_2, ..., n_m \}$ to $E^1_k$, and the data from neighboring nodes of class $l$ to $E^2_{k, l}$. In contrast to \cite{Jain:2016}, we use a weighted sum pooling for the edge inputs, i.e., $X'_i = \sum_{j, cl(j)=k}w_{i,j}X_j$. This pooling has shown to be more suitable for ST data forecasting. Finally, the output from the two RNNs are concatenated and fed to a node RNN $N^1_j$, which then predicts the number of events at time $t$. For each epoch during training, we can also sample the nodes to maintain scalability when dealing with large graphs. The sampling can be done non-uniformly across groups, e.g., sampling more often groups that contribute higher to the overall error.

%\xnote{\bf Pseudo description of the SRNN algorithm.}

\begin{algorithm}
\caption{GSRNN for ST Forecasting.}\label{alg:SRNN}
\begin{algorithmic}
    \State \textbf{Input:} Input crime intensity $\{x_i(t)\}_{t=1}^n$,  for all nodes $i = 1\dots N$.
    \State \textbf{Output:} Predicted crime intensity $x_{t}(i)$ at time slot $t = n+1$ for all nodes $i$.
    \State \textbf{Step 1:}  Infer the mutual excitation coefficient using the Hawkes model $w_{ij}$ for the multivariate time series $x_t(i)$, and set as graph weights.
    \State \textbf{Step 2:} Partition the nodes to $K$ classes according to total crime count.
    \State \textbf{Step 3:} Preprocess each time series $x_t(i)$ by apply superresolution and integration as in Eqns .(\ref{Map}) and (\ref{TS}).%Algorithm \ref{SingleNodeDNN}.
    \State \textbf{Step 4:} Construct GSRNN model where the edge RNN outputs are pooled via a weighted sum $\sum_{cl(j) = c}w_{ij} x_{j}$.
    \State \textbf{Step 5:} Train network via ADAM,  optionally subsample the nodes in each class for efficiency.
    \State \textbf{Step 6:} Apply the inverse maps to the data augmentation to recover the predicted crime intensity at the time $n+1$.
\end{algorithmic}
\end{algorithm}
\vskip -1cm

%==============================================================================
\section{ST Crime Forecasting Results}
\label{ExperimentsCrime}
We compare two naive strategies that do not utilize the STWG %spatial graph
information against the GSRNN %Bidirectional SRNN
model. The first, denoted by Single Node, trains a separate LSTM model on each individual zip code. The second, denoted as Joint Training, organizes the zip code regions in three groups according to the average crime rate  (Group1 contains the zip code regions with lowest crime rate, and so forth.). The RNN is trained jointly for each group.

To construct the STWG used in the GSRNN model,  a K-nearest neighbor graph of $K = 15$ is used as the initial sparse structure for the MHP inference algorithm. The obtained self excitation rates $a_{ij}$ are further thresholded to reach a sparsity rate of 0.1, and then normalized. Namely, $w_{ij} = \frac{a_{ij}}{max(a_{ij})}$.

For both single node and joint training, a 2-layer LSTM with 128 and 64 units is used,
where a dropout rate of 0.2 is applied to the output of each layer. For the GSRNN model, a 64/128 unit single layer LSTM is used for
the edge/node RNN, respectively.

All models are trained using the ADAM optimizer with a learning rate of 0.001 and other default parameters.  %$\alpha_1 = 0.9$, $\beta_1 = 0.999$.
We compare the RMSE in both CDF (diurnal cumulated time series) and PDF (raw time series) of the predictions. For the LA Data, we test on the last two months, and use the rest for training. For CHI we test on the last one month (31 days), and use the rest for training.  See Tables 3 and 4.
%The model's parameters are as follows.
%\begin{table}[!t]
%\caption{Average RMSE on for LA Crime Data. Left: RMSE on CDF, Right: RMSE on PDF. }
%\renewcommand{\arraystretch}{1.3}
%\centering
%\begin{tabular}{|c|c|c|c|}\hline
%  & Single Node & Joint Training & Bidirectional SRNN \\ \hline
%Group 1&0.1076/0.1128   &0.0623/0.0754    &0.0589/0.0775    \\\hline
%Group 2&0.1536/0.1649   &0.1017/0.1242    &0.0815/0.1094    \\\hline
%Group 3&0.2351/0.2507   &0.1676/0.1913    &0.1436/0.1834    \\\hline
%Average &0.1736/0.1851  &0.1204/0.1401    &0.1019/0.1311    \\\hline
%\end{tabular}
%\end{table}
\begin{table}[!t]
\caption{Average RMSE on for LA Crime Data. Left: RMSE on CDF, Right: RMSE on PDF. }
\renewcommand{\arraystretch}{1.3}
\centering
\begin{tabular}{|c|c|c|c|}\hline
  & Single Node & Joint Training & GSRNN \\ \hline
Group 1&0.108/0.113   &0.062/0.075    &0.059/0.078    \\\hline
Group 2&0.154/0.165   &0.102/0.124    &0.082/0.109    \\\hline
Group 3&0.235/0.251   &0.168/0.191    &0.144/0.183    \\\hline
Average &0.174/0.185  &0.120/0.140    &0.103/0.131    \\\hline
\end{tabular}
\end{table}
%\begin{table}[!t]
%\caption{Average RMSE on for Los Angeles Crime Data. Top Row: RMSE on CDF, Bottom Row: RMSE on PDF. }
%\renewcommand{\arraystretch}{1.3}
%\centering
%\begin{tabular}{|c|c|c|c|}\hline
%  & Single Node & Joint Training & Bidirectional SRNN \\ \hline
%Group 1&0.1076   &0.0623    &0.0589    \\
%       &0.1128   &0.0754    &0.0775    \\\hline
%Group 2&0.1536  &0.1017   &0.0815      \\
%       &0.1649   &0.1242    &0.1094      \\\hline
%Group 3&0.2351   &0.1676   &0.1436      \\
%       &0.2507   &0.1913    &0.1834      \\\hline
%Average &0.1736   &0.1204   & 0.1019    \\
%        &0.1851   &0.1401   & 0.1311     \\ \hline
%\end{tabular}
%\end{table}
%\begin{table}[!t]
%\caption{Average RMSE on for CHI Crime Data. Left: RMSE on CDF, Right: RMSE on PDF. }
%\renewcommand{\arraystretch}{1.3}
%\centering
%\begin{tabular}{|c|c|c|c|}\hline
%  & Single Node & Joint Training & Bidirectional SRNN \\ \hline
%Group 1&0.1741/0.1662   &0.2039/0.1419    &0.1020/0.1084     \\\hline
%Group 2&0.3807/0.3475   &0.1519/0.1315   &0.1810/0.1965      \\\hline
%Group 3&0.6999/0.6965   &0.4127/0.4954   &0.3823/0.4124      \\\hline
%Average &0.4821/0.4713   &0.2863/0.3167   &0.2577/0.2781     \\\hline
%\end{tabular}
%\end{table}
\begin{table}[!t]
%\caption{Average RMSE on for CHI Crime Data. Left: RMSE on CDF, Right: RMSE on PDF. All statistics are associated with unit: number of crimes.}
\caption{Average RMSE on for CHI Crime Data. Left: RMSE on CDF, Right: RMSE on PDF. Unit: number of crimes.}
\renewcommand{\arraystretch}{1.3}
\centering
\begin{tabular}{|c|c|c|c|}\hline
  & Single Node & Joint Training & GSRNN \\ \hline
Group 1&0.174/0.166   &0.204/0.143   &0.102/0.108     \\\hline
Group 2&0.381/0.348   &0.153/0.133   &0.181/0.197      \\\hline
Group 3&0.699/0.697   &0.413/0.495   &0.382/0.412      \\\hline
Average &0.482/0.471  &0.286/0.317   &0.258/0.278     \\\hline
\end{tabular}
\end{table}
%\begin{table}[!t]
%\caption{Average RMSE on for Chicago Crime Data. Top Row: RMSE on CDF, Bottom Row: RMSE on PDF. }
%\renewcommand{\arraystretch}{1.3}
%\centering
%\begin{tabular}{|c|c|c|c|}\hline
%  & Single Node & Joint Training & Bidirectional SRNN \\ \hline
%Group 1&0.1741   &0.2039    &0.1020     \\
%       &0.1662   &0.1419    &0.1084      \\\hline
%Group 2&0.3807   &0.1519   &0.1810      \\
%       &0.3475   &0.1315    &0.1965      \\\hline
%Group 3&0.6999   &0.4127   &0.3823      \\
%       &0.6965   &0.4954   &0.4124      \\\hline
%Average &0.4821   &0.2863   &0.2577     \\
%        &0.4713   &0.3167   &0.2781    \\ \hline
%\end{tabular}
%\end{table}

\begin{figure}
\centering
\begin{tabular}{cc}
\includegraphics[width=0.48\columnwidth]{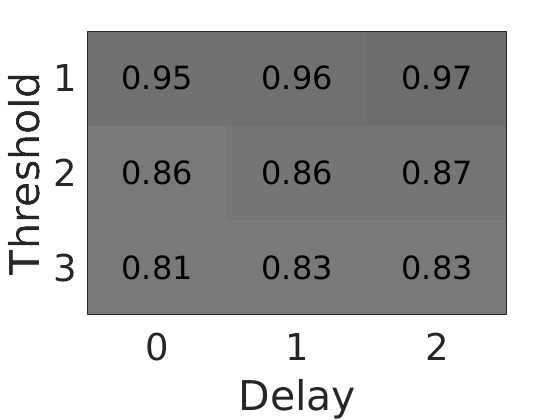}&
\hskip -0.5cm
\includegraphics[width=0.48\columnwidth]{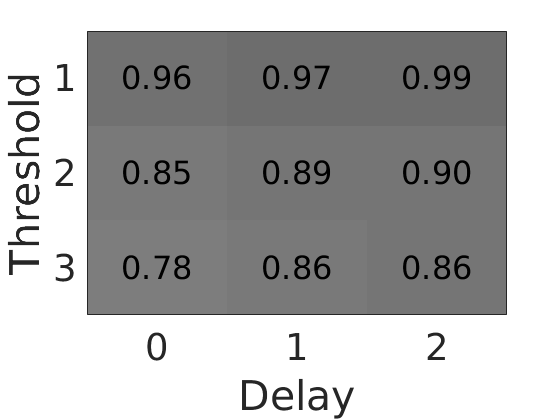}\\
(a) Los Angeles &(b) Chicago\\
\end{tabular}
%\caption{ \xmod{Precision matrix of GSRNN for Los Angeles and Chicago averaged across top all zipcodes with at least one hourly time slot containing more than 3 crimes. } }
\caption{Precision matrix of GSRNN for Los Angeles and Chicago averaged across top all zipcodes with at least one hourly time slot containing more than 3 crimes.}
\label{precision-matrix-all}
\end{figure}

\begin{figure}
\centering
\begin{tabular}{cc}
\includegraphics[width=0.44\columnwidth]{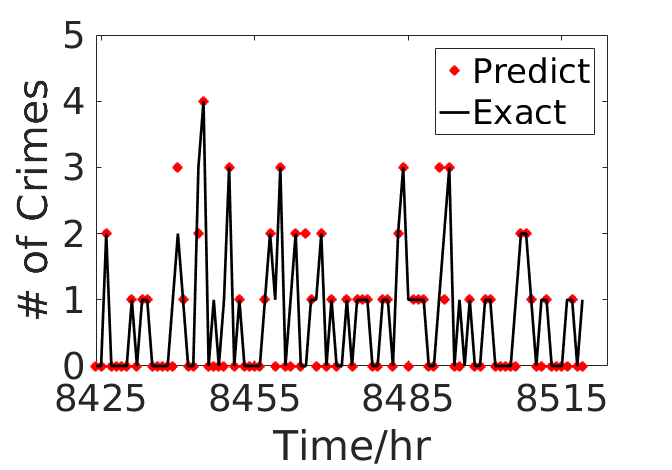}&
\includegraphics[width=0.44\columnwidth]{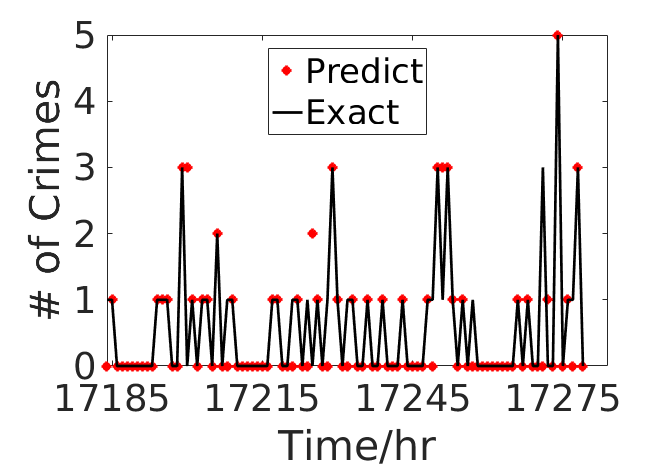}\\
(a)CHI zip code 60171&(b)LA zip code 90003\\
\end{tabular}
\caption{Panels a), b) plot snapshots of the predicted vs exact hourly crime rate for CHI and LA data.}
\label{fig:crimeNode}
\end{figure}
We observe that Joint Training leads to a performance boost compared to the Single Node approach, and adding the bidirectional graph leads to a further performance increase. Theses conclusions are consistent across the stratified groups as well. The precision matrices (thresholded to three in both number of crimes and delay) averaged over all the nodes for LA and CHI are plotted in Fig. \ref{precision-matrix-all}, respectively. Figure \ref{fig:crimeNode} shows the predicted and exact crime time series over two graph nodes.

%==============================================================================
\section{ST Traffic Forecasting Results}
\label{ExperimentsTraffic}
%\xnote{Feel free to restructure below if you see fit. }

% \subsection{stacked LSTM for Nodewise Traffic Forecasting}
% \subsection{stacked LSTM for Traffic Forecasting on kNN Graph}
% \subsection{spatio-Temporal Structured LSTM for Crime Forecasting on kNN Graph}
% \subsection{Comparison Between Different Methods}

We benchmark our methodology on two public datasets for traffic forecasting \cite{Junbo:2017}.  The BikeNYC
%dataset contains coordinates from the NYC bike system from April 1st to September 30th 2014.
and TaxiBJ datasets are both embedded in rectangular bounding boxes,
forming a rectangular grid of size 32 x 32 and 16 x 8 respectively.
%(see Fig.\ref{fig:ny2dl} and Fig.\ref{fig:bj2d}).

%contains GPS data for taxi cabs in Beijing from 4 time intervals from the year 2014 to 2016: 1st July 2013 - 30th October. 2013,
%1st March 2014 - 30th June 2014, 1st March 2015 - 30th June 2015, 1st Nov. 2015 - 10th April 2016. The GPS coordinates from both datasets are
%embedded in a rectangular bounding box, forming a rectangular grid of size 32 x 32 and 16 x 8 respectively(see Fig.\ref{fig:bjcl} and Fig.\ref{fig:ny2dl}).

\begin{figure}
\centering
\begin{tabular}{ccc}
\includegraphics[height=0.25\columnwidth, width=0.3\columnwidth]{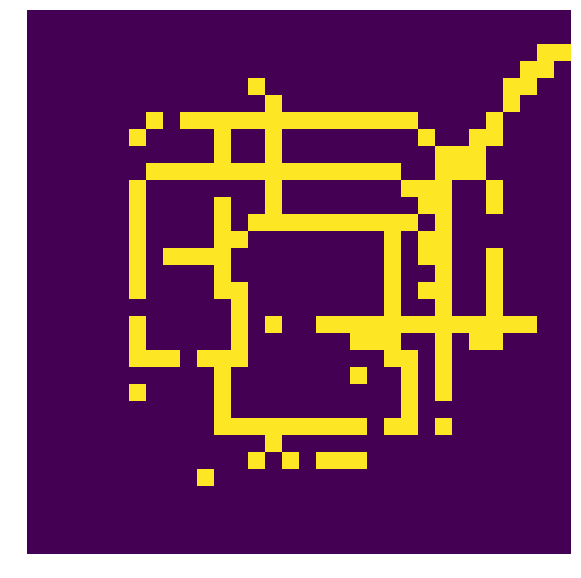}&
\includegraphics[height=0.25\columnwidth, width=0.3\columnwidth]{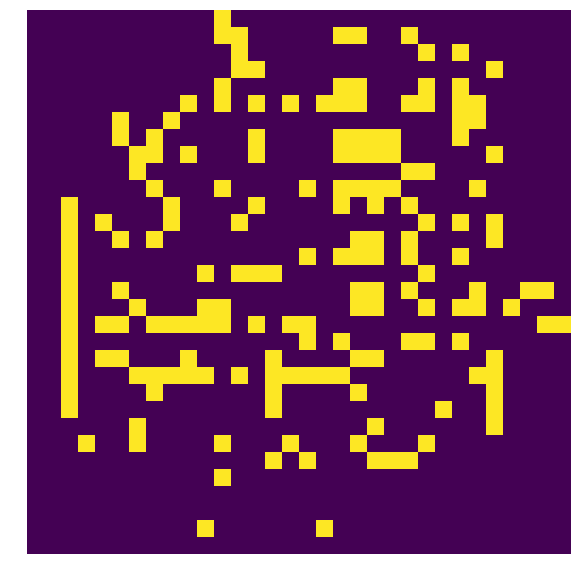}&
\includegraphics[height=0.25\columnwidth, width=0.3\columnwidth]{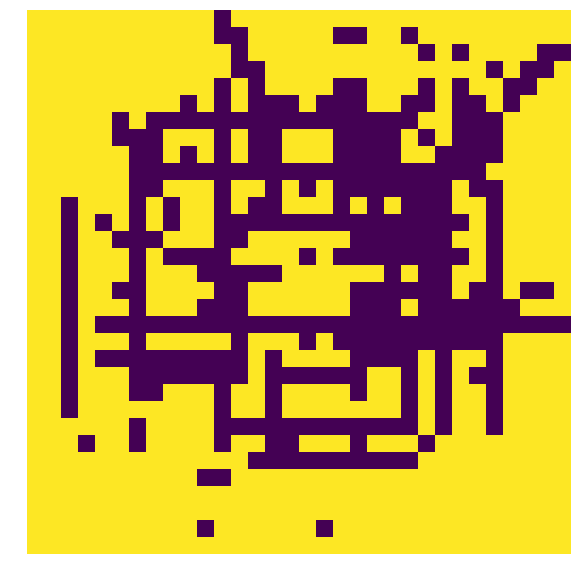}\\
(a)&(b)&(c)\\
\end{tabular}
\caption{Panels a)-c) visualize the node class assignment from group 1 - 3 in the Beijing Traffic data respectively, where a yellow pixel indicates the assignment of the pixel node to its corresponding class. For example, the yellow pixels in panel a) are grouped to class 1. }
\label{fig:bjcl2}
\end{figure}

To cast the problem into the graph representation %graph temporal
framework used in this paper, we consider each pixel in the spatial grid as a graph node, and connect each node with its four immediate neighbors. The graph weights are set to $1/4$ for all edges, the same as in an unweighted graph. More sophisticated methods could be used for graph construction, but we found the 4-regular graph already yields good performance.
The nodes are then sorted into three classes according to the overall cumulative traffic count. For the New York data, there are three equal size classes, whereas in the Beijing dataset, the classification is picked manually to reflect the geographical structure of the Beijing road system (see Fig. \ref{fig:bjcl2}).

For the BikeNYC, we use a two layer LSTM with (32, 64) units and 0.2 dropout rate at each layer for the single-node model, and a two layer LSTM with (64, 128) units and 0.2 dropout rate at each layer for the joint and GSRNN model. For the TaxiBJ, we use a two layer LSTM with (64,128) units for the single-node, and a three layer LSTM model with (64, 128, 64) units for the joint model; for the GSRNN model, the edge RNN is a two layer LSTM with (64, 128) units, and the node RNN is a one layer LSTM with 128 units.
All models are trained using the ADAM optimizer with a learning rate of 0.001 and other default parameters. %$\alpha_1 = 0.9$, $\beta_1 = 0.999$.
The learning rate is halved every 50 epochs, and a total of 500 epochs is used for training.

\begin{figure}
\centering
\begin{tabular}{cc}
\includegraphics[width=0.48\columnwidth]{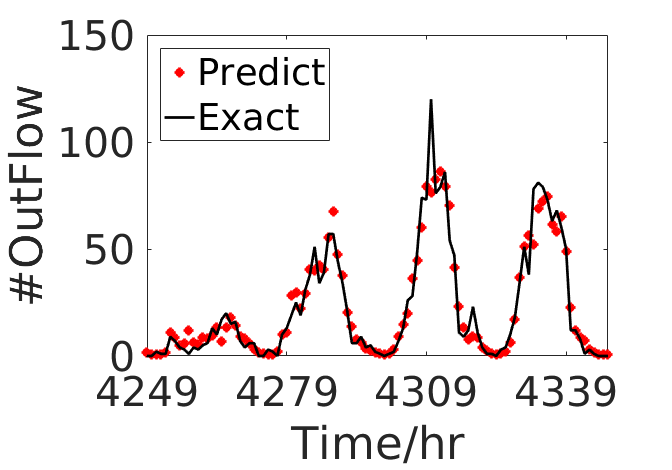}&
\hskip -0.6cm
\includegraphics[width=0.48\columnwidth]{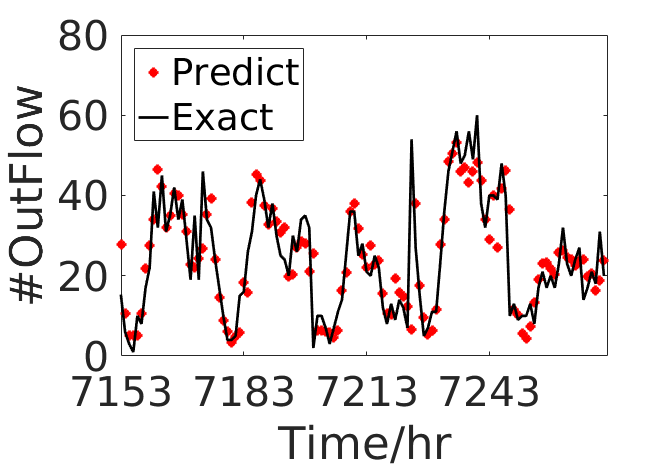}\\
New York: $x= (5,5)$  &Beijing: $x=(12,7)$\\
\end{tabular}
\caption{Comparison
between predicted vs exact traffic out-flow at a specified point $x$ for New York and Beijing. }
\label{fig:bjny}
\end{figure}

\begin{figure}
\centering
\begin{tabular}{ccc}
\hskip -0.7cm
\includegraphics[width=0.37\columnwidth]{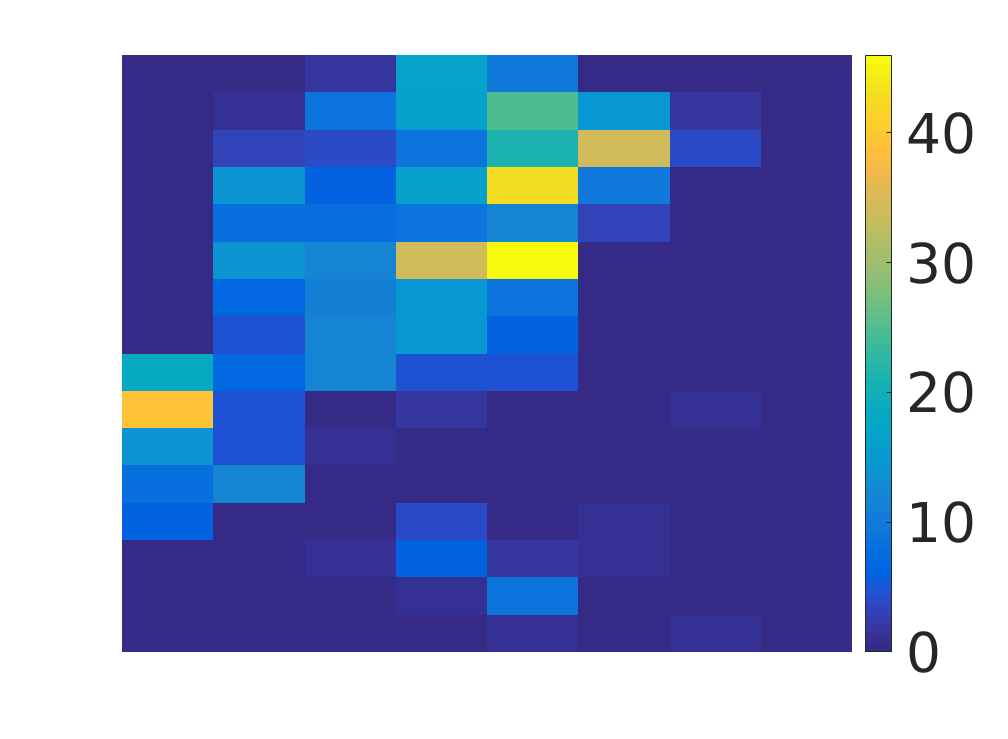}&
\hskip -0.4cm
\includegraphics[width=0.37\columnwidth]{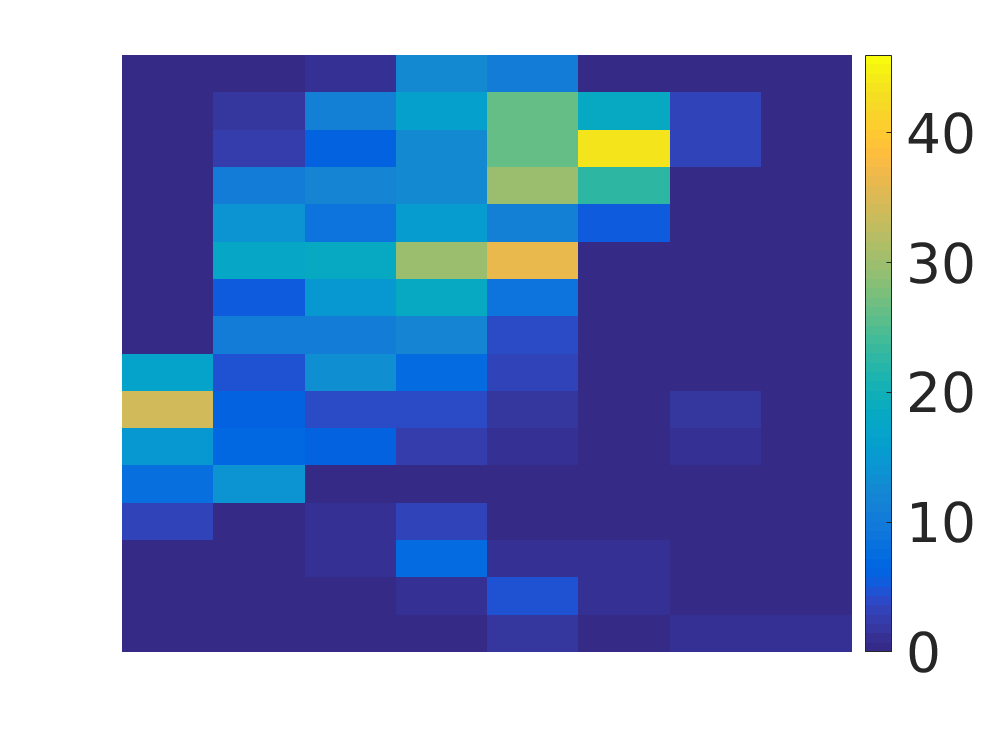}&
\hskip -0.4cm
\includegraphics[width=0.37\columnwidth]{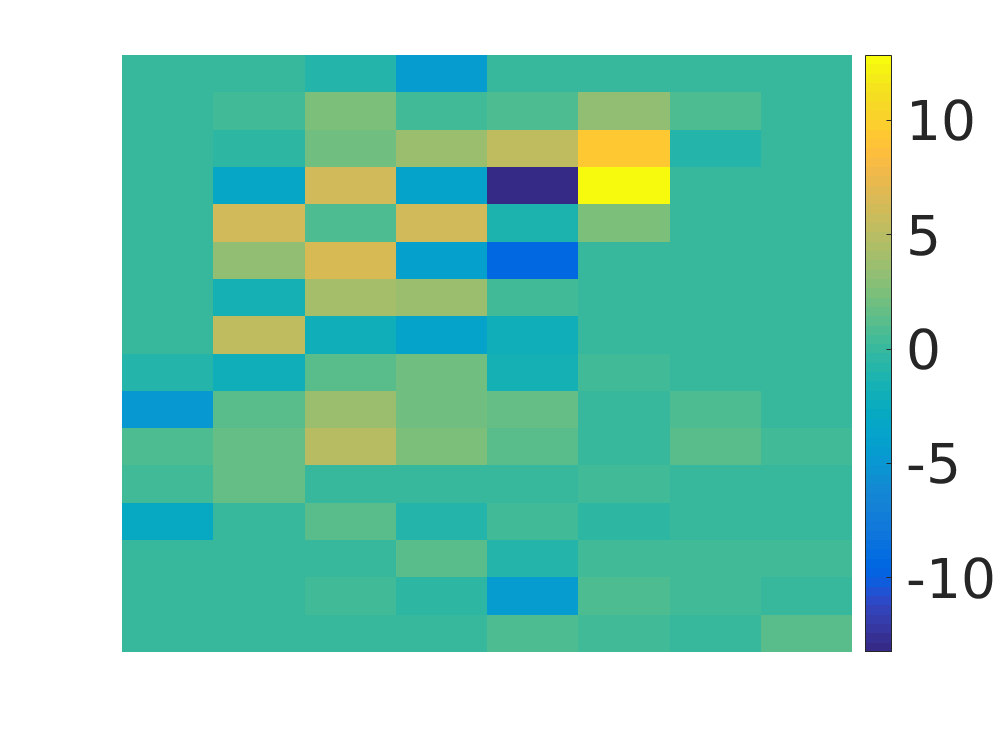}\\
(a) predicted &(b) exact &(c) difference\\
\end{tabular}
\caption{Comparison between predicted (a) and exact (b) traffic out-flow at $t=8647$ for New York city over a $16\times 8$ grid. Difference shown in c).}
\label{fig:ny2dl}
\end{figure}

\begin{figure}
\centering
\begin{tabular}{ccc}
\hskip -0.7cm
\includegraphics[width=0.37\columnwidth]{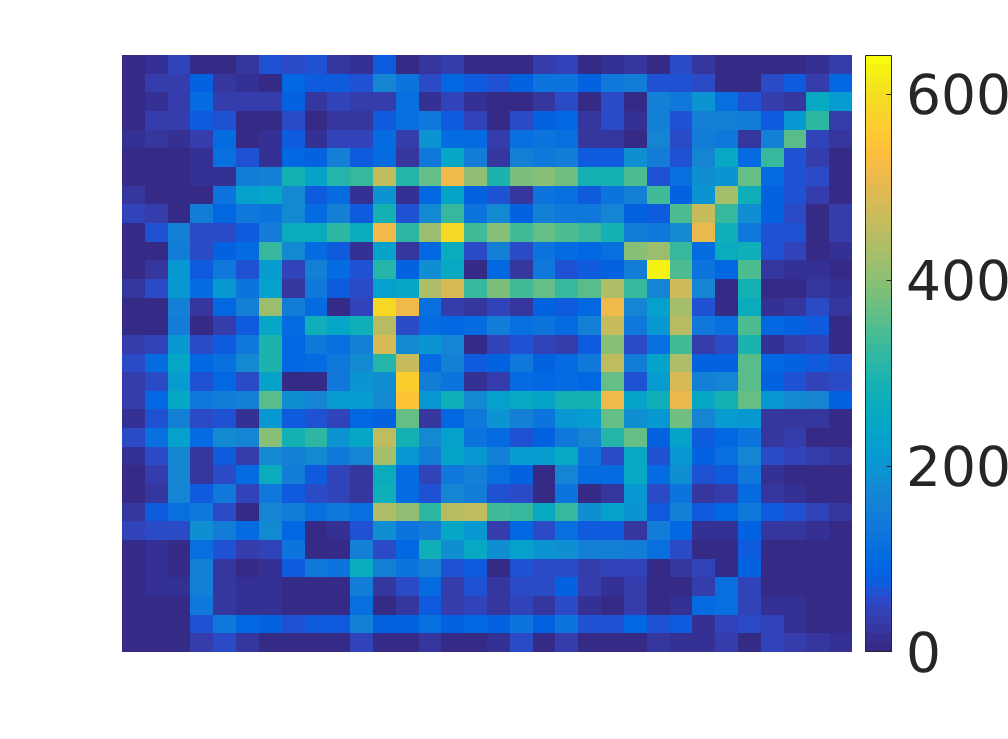}&
\hskip -0.3cm
\includegraphics[width=0.37\columnwidth]{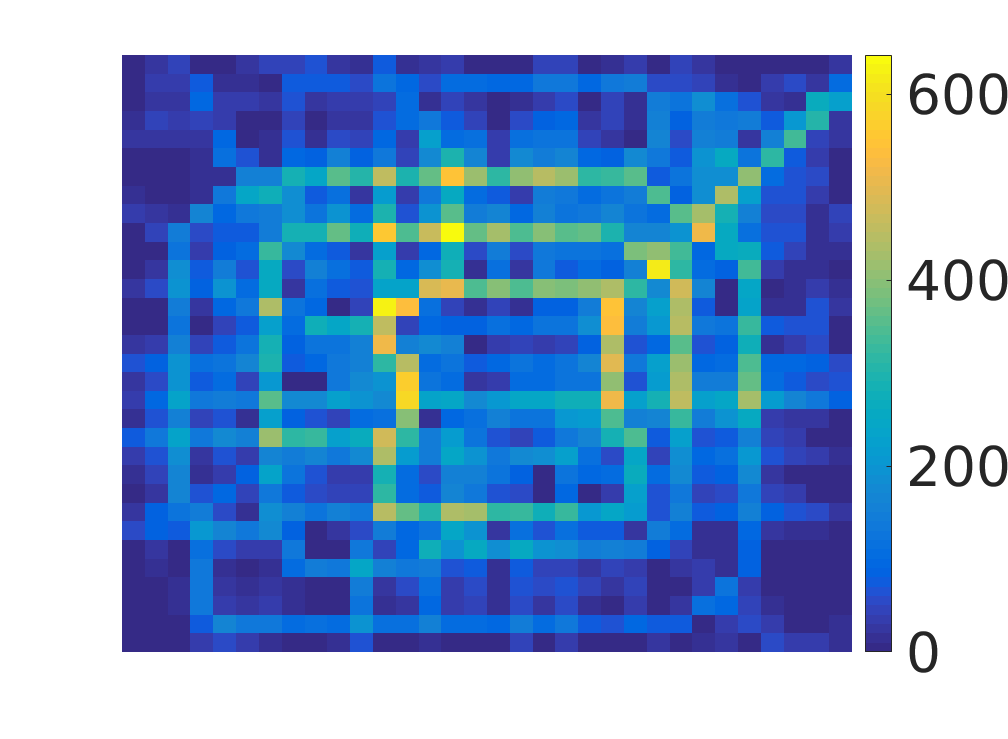}&
\hskip -0.3cm
\includegraphics[width=0.37\columnwidth]{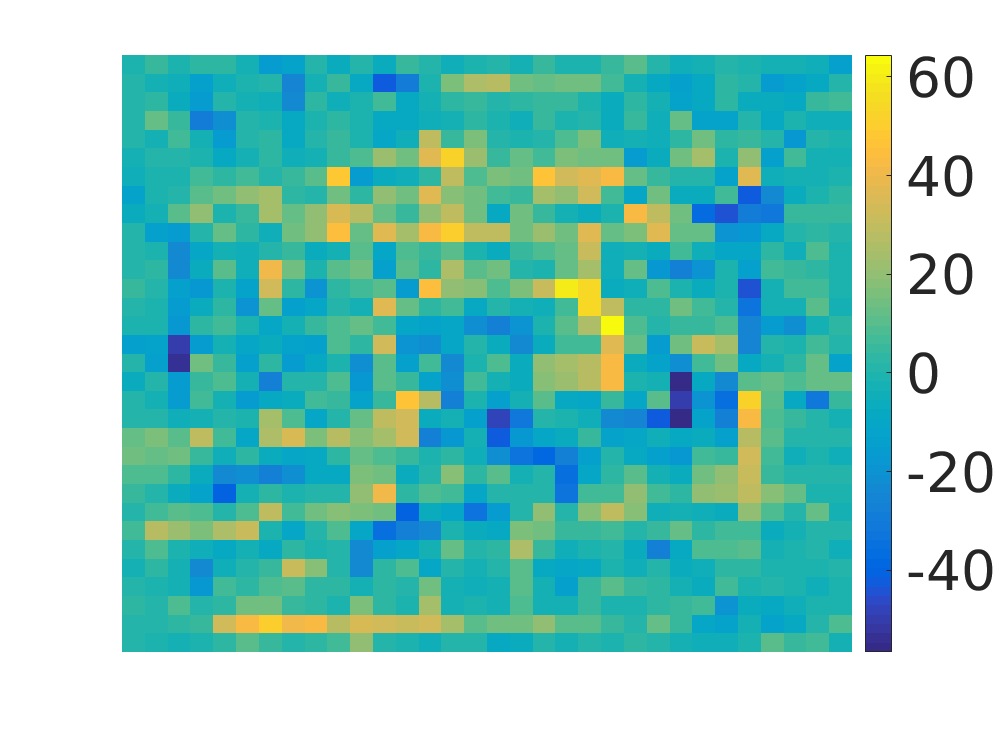}\\
(a) predicted &(b) exact &(c) difference\\
\end{tabular}
\caption{Comparison between predicted (a) and exact (b) traffic out-flow at $t = 17204$ for Beijing over a $32\times 32$ grid. Difference shown in (c).}
\label{fig:bj2d}
\end{figure}

For evaluation, we use the Root Mean Square Error (RMSE) across all nodes and all time intervals. The same train-test split is used in our experiments as in \cite{Junbo:2017}. The results on RMSE (Table \ref{Traffic:Compare}) are reported on the testing error based on the model parameters with the best validation loss. Comparisons between the predicted and exact traffic on two grids over a randomly selected time period is shown in Fig. \ref{fig:bjny}. On a randomly selected time slot, we plot the predicted and exact spatial data and errors in Figs. \ref{fig:ny2dl} and \ref{fig:bj2d}.

%\begin{table}[!h]
%\caption{RMSE on for Traffic Data. }
%\renewcommand{\arraystretch}{1.3}
%\centering
%\begin{tabular}{|c|c|c|c|c|}\hline
%  & \tiny{Single Node} & \tiny{Joint Training} & \tiny{Bidirectional SRNN} & \tiny{STResNet \cite{Junbo:2017}} \\ \hline
%Beijing &23.50   &19.49   & {\bf 16.59} & 16.69  \\ \hline
%NY &6.7738  &6.332   & {\bf 6.0854} & 6.33   \\  \hline
%\end{tabular}
%\end{table}

\begin{table}[!h]
\caption{RMSE on for Traffic Data. }
\renewcommand{\arraystretch}{1.3}
\centering
\begin{tabular}{|c|c|c|c|c|}\hline
  & {Single} &{Joint} & {GSRNN} & {STResNet } \\
  &{Node}&{Training}&{}&\cite{Junbo:2017}\\
  \hline
Beijing &23.50   &19.5   & {\bf 16.59} & 16.69  \\ \hline
NY &6.77         &6.33   & {\bf 6.08}  & 6.33   \\  \hline
\end{tabular}
\label{Traffic:Compare}
\end{table}

\section{Conclusion}
\label{Conclusion}
% 1. Summary of the contribution

% 2. Advantages
%In this paper, we studied the spatio-temporal (ST) data, especially the crime data, comprehensively.
We develop a multiscale framework that contains two components: inference of the macroscale spatial temporal graph representation of the data, and a generalizeable graph-structured recurrent neural network (GSRNN) to approximate the time series on the STWG. Our GSRNN is arranged like a feed forward multilayer perceptron with each node and each edge associated with LSTM cascades instead of weights and activation functions. To reduce the model's complexity, we apply weight sharing among certain type of edges and nodes. This specially designed deep neural network (DNN) takes advantage of the RNN's ability to learn the pattern of time series, capturing real time interactions of each node to its connected neighbors. To predict the value of the time series for a node at the next time step, we use the information of its neighbors and real time interactions. For the ST sparse data, we propose efficient data augmentation techniques to boost the DNN's performance. Our model demonstrates remarkable results on both crime and traffic data; for crime data we measure the performance with both root mean squared error (RMSE) and the proposed precision matrix.

The method developed here forecasts crime on the time scale of an hour in each US zip code region.  This is in contrast to the commercial software PredPol (www.predpol.com) that forecasts on a smaller spatial scale and longer timescale.  Due to the different scales, the methods have different uses - PredPol is used to target locations for patrol cars to disrupt crime whereas the method proposed here might be used for resource allocation on an hourly basis within different patrol regions.

There are a few issues that require future attention. The space on which the data is distributed is represented as a static graph. A dynamic graph that better models the changing mutual influence between neighboring nodes could be incorporated in our framework. Furthermore, in the traffic forecasting problem, better spatial representation of the traffic data could also be explored. Our model could also be applied to broader fields, e.g., quantitative finance and social networks \cite{Zipkin:2016}.

\section{Acknowledgments}
This work was supported by ONR grant N00014-16-1-2119 and NSF DMS grants 1417674 and 1737770. The authors thank the Los Angeles Police Department for providing the crime data. We thank Dr. Da Kuang for help with the QGIS software.

\end{document}